\pdfoutput=1
\RequirePackage{fix-cm}
\documentclass[11pt,letterpaper,logo,onecolumn,copyright]{microsoft}
\usepackage{ms-colors}

\usepackage{natbib}
\usepackage{xurl}
\usepackage{array}
\usepackage{multirow}
\newcommand{\sys}{LLM-as-Jev}
\newcommand{\cfgIS}{Intent-100}
\newcommand{\cfgIL}{Intent-1ep}
\newcommand{\cfgRS}{Reason-9k}
\newcommand{\cfgRM}{Reason-21k}
\newcommand{\cfgRL}{Reason-30k}
\newcommand{\cfgLong}{Reason+Long-12k}
\newcommand{\twoline}[2]{\begin{tabular}[b]{@{}c@{}}#1\\#2\end{tabular}}

\newcommand{\tk}[1]{{\setlength{\fboxsep}{1pt}\fbox{\texttt{\strut #1}}}}

\newcommand{\colfigwidth}{0.6\linewidth}

\hypersetup{
  pdftitle={LLM-as-Jev: LLMs Are Already Jev-Style Decision Models\textemdash When and How to Fine-Tune Them},
  pdfauthor={Yinheng Li and Justin Wagle}
}

\title{\sys{}: LLMs Are Already Jev-Style Decision Models---When and How to Fine-Tune Them}
\author{Yinheng Li and Justin Wagle}
\correspondingauthor{\email{yinhengli@microsoft.com}}

\begin{document}

\begin{abstract}
Jev-style decision models have gained surging traction by making decision-making dramatically faster and more efficient, returning categorical probability distributions over predefined options without generating free-form text. However, because Jev's architecture and training methodologies remain proprietary, how to replicate these capabilities effectively remains an open question. In this work, we investigate the extent to which general-purpose LLMs already possess this capability out of the box, and when fine-tuning is actually necessary. We present \sys{}, an architecture-preserving framework offering both a training-free inference recipe and a fine-tuning objective with KL-divergence anchors. Across Qwen3.5-4B and Qwen3-0.6B, we find that modern LLMs are inherently effective decision models: off the shelf, frozen models match or exceed existing training-free readouts on the same 4B backbone, scale to arbitrary option counts, and natively handle multimodal decisions. They also seamlessly inherit the backbone's thinking mode: while Jev was introduced as a System~1 model, ``Thinking Jev'' leverages budgeted test-time reasoning to boost accuracy on challenging tasks without training. Fine-tuning provides targeted gains for smaller models and complex routing, while our KL anchors prevent behavioral degradation in general text generation.
\end{abstract}

\maketitle
\suppressfloats[t]

\section{Introduction}
\label{sec:intro}
\begin{figure*}[t]
\centering
\includegraphics[width=\textwidth]{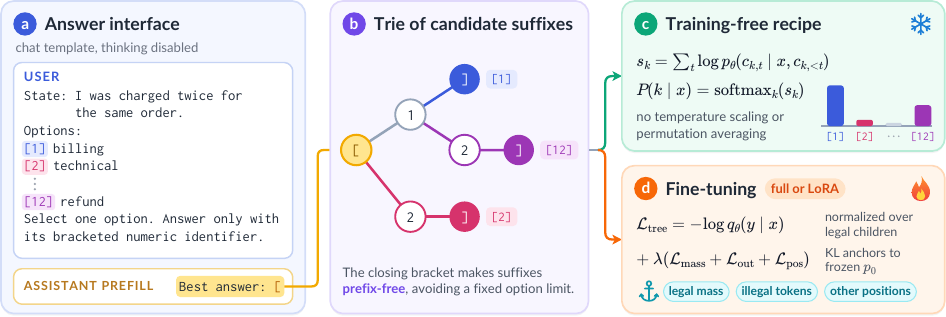}
\caption{Overview of \sys{}, illustrating prompt prefilling, candidate suffix trie, joint training-free scoring, and anchored fine-tuning.}
\label{fig:framework}
\end{figure*}

Software systems routinely make decisions over predefined options: routing support tickets, classifying intent, or verifying evidence rules. While chat models can perform these tasks, parsing generated text introduces latency and formatting failures. Jev~\citep{typesafe2026jev} addresses this by directly returning a probability distribution across candidate options without free-form generation. Jev aims to be fast and cost-effective while retaining LLM versatility: deploying a new task requires only a prompt and option list rather than a dedicated classifier. Although Jev's weights remain proprietary, the public JevBench leaderboard~\citep{benchmarkheaven2026jevbench} already lists over 60 community Jev-style models.

A central design choice among community Jev-style models is whether to update model parameters. Many approaches fine-tune the backbone with adapters or classification heads; others evaluate frozen LLMs directly. Surprisingly, fine-tuning does not guarantee superior performance: for instance, reflex 4B~\citep{reflex} achieves 79.7\% accuracy on public JevBench items, slightly trailing SemIf~\citep{lee2026semif}, which simply extracts option-letter logits from the unmodified frozen model (80.5\%; Table~\ref{tab:training-free}). This raises two fundamental questions: \emph{to what extent do pre-trained LLMs inherently possess decision-making capabilities, and under what conditions does fine-tuning yield practical gains?}

\paragraph{Our approach: leveraging off-the-shelf LLMs.}
We present a framework that turns existing causal LLMs into Jev-style decision models without altering their architectures, tokenizers, or vocabularies, while minimizing dependence on task-specific training. In contrast to many community Jev-style models that introduce custom classification heads or reserved tokens~\citep{kev,imajev,decisionfast}---which must be retrained for every new backbone---our architecture-preserving design provides three distinct benefits:
\begin{itemize}
\item \textbf{Zero architectural overhead per backbone.} The same prompting convention and parallel scoring procedure apply out of the box to any causal LLM, allowing newly released models to be evaluated immediately and fine-tuned only when targeted adaptation is required.
\item \textbf{Direct inheritance of base model scaling.} As foundation models advance in general reasoning and factual knowledge, their decision-making accuracy improves correspondingly without necessitating new decision-specific training corpora.
\item \textbf{Native multimodal and reasoning support.} Multimodal backbones evaluate visual queries through the identical interface without multimodal tuning (\S\ref{sec:training-free}). Furthermore, by enabling the model's native thinking mode under a hard token budget, \sys{} unlocks ``Thinking Jev'' to tackle complex decisions with controllable inference costs (\S\ref{sec:thinking}).
\end{itemize}

\paragraph{Two complementary recipes.}
As illustrated in Figure~\ref{fig:framework}, \sys{} offers two recipes under a unified interface: a \textbf{training-free recipe} that extracts decisions directly from prefix-conditioned next-token probabilities without parameter updates, and a \textbf{fine-tuning recipe} that optimizes candidate discrimination via listwise loss while employing KL anchors to preserve base model capabilities.

\paragraph{Findings.}
We evaluate both recipes on Qwen3.5-4B and Qwen3-0.6B across extensive benchmarks:
\begin{enumerate}
\item Pre-trained LLMs are inherently effective decision models out of the box. Without training, frozen Qwen3.5-4B matches or exceeds existing training-free readouts on the same backbone, scales to arbitrary option counts, and natively handles multimodal visual decisions (\S\ref{sec:training-free}).
\item Fine-tuning provides targeted rather than universal gains. It substantially benefits compact models and specialized domains (such as high-cardinality routing), but yields diminishing returns for capable backbones on broad benchmarks (\S\ref{sec:finetuning}).
\item Parameter adaptation requires explicit regularization to prevent behavioral drift. Pairing parameter-efficient LoRA with KL divergence anchors prevents generative collapse while achieving our best non-thinking 4B accuracy (\S\ref{sec:recipe}).
\item Unlocking native thinking under an enforceable token budget allows the frozen 4B model to resolve complex multi-hop queries, reaching 94.8\% accuracy on JevBench at strictly controllable inference costs (\S\ref{sec:thinking}).
\end{enumerate}

\section{Background and Related Work}
\label{sec:related}

\paragraph{Decision models, Jev, and JevBench.}
Traditionally, classification tasks relied on discriminative models like BERT~\citep{devlin2019bert}, which require dedicated heads and supervised fine-tuning over fixed label sets. In contrast, Jev~\citep{typesafe2026jev} functions like a general LLM: it evaluates arbitrary tasks zero-shot through prompting, directly emitting categorical probability distributions over discrete candidates without text generation. Jev formalizes decisions across three formats: \emph{choice} ($K$ options), \emph{noul} (binary verification), and \emph{score} (ordinal grading). To assess such systems, JevBench~\citep{benchmarkheaven2026jevbench} benchmarks decision models across intelligence, calibration, latency, and cost across 231 public items (easy, standard, hard), while sealed splits remain private.

\paragraph{Community Jev-style models.}
Table~\ref{tab:community} (Appendix~\ref{app:community}) summarizes community efforts, which broadly fall into two paradigms:
\emph{Frozen-model readouts} extract predictions from output logits without weight updates, using option-letter logits~\citep{lee2026semif,openalternativejev}, per-option querying~\citep{yinsongxu2026llm2jev}, label-prior and option-order correction~\citep{zhang2026anyjev}, or temperature-scaled letter logits on larger backbones~\citep{cygnet,jqv}. In contrast, \sys{} jointly scores arbitrary candidates in a single prompt (\S\ref{sec:scoring}), matching or outperforming the same-backbone frozen readouts in Table~\ref{tab:community} (\S\ref{sec:training-free}).
\emph{Fine-tuned readouts} adapt backbones via LoRA or full tuning on synthetic or replayed data~\citep{jevk5,plumb,hopper,reflex,decider}. However, they typically rely on costly teacher-generated data and risk representation drift, whereas our anchored objective preserves general model capabilities.

\paragraph{Probabilistic calibration.}
In automated decision systems, well-calibrated probabilities are as vital as top-1 accuracy, allowing downstream workflows to defer low-confidence predictions to human review. Because modern neural networks can be poorly calibrated~\citep{guo2017calibration} and LLM calibration varies with scale and fine-tuning~\citep{desai2020calibration,kadavath2022language,tian2023just}, we evaluate decision reliability using Expected Calibration Error (ECE) and Brier score directly on raw model distributions without post-hoc temperature scaling.

\section{The \sys{} Framework}
\label{sec:method}

\subsection{Problem Formulation}
A decision instance consists of a context state $s$, a set of candidate options $o_1,\dots,o_K$, and a target choice ${y \in \{1,\dots,K\}}$. The objective is to produce a probability distribution ${P(k\mid s)}$ over the candidate set. All three Jev-supported task types map cleanly into this formulation: a verification (yes/no) query specifies two options describing the acceptance criteria for ``yes'' and ``no'', while an ordinal scoring query defines an ordered sequence of graded options.

\subsection{Answer Interface}
\label{sec:interface}
We query the model using its default chat template with thinking mode disabled, formatting the user prompt as follows:
\begin{quote}\small\ttfamily
State: \{$s$\}\\
Options: [1] \{$o_1$\}\ \dots\ [K] \{$o_K$\}\\
Select one option. Answer only with its bracketed numeric identifier.
\end{quote}
Each candidate is rendered on a separate line. As shown in Figure~\ref{fig:framework}(a), we condition the assistant's turn on the fixed prefix \texttt{Best answer: [}, such that selecting option $k$ corresponds to generating the completion ``$k$\texttt{]}''. We define this completion as the \emph{candidate suffix} (Figure~\ref{fig:framework}b), represented by the token sequence $c_k=(c_{k,1},\dots,c_{k,n_k})$. This design yields three key advantages:
\begin{itemize}
\item \textbf{Scaling to arbitrary options.} Because every candidate suffix terminates with a closing bracket, no complete suffix can serve as the prefix of another (e.g., \texttt{1]} is not a prefix of \texttt{12]}). This prefix-free property supports arbitrarily large candidate sets $K$, unlike single-letter readouts that are inherently bounded to 16 or 26 options.
\item \textbf{Fast parallel evaluation.} The candidate suffixes naturally form a shallow prefix tree (Figure~\ref{fig:framework}b) whose depth is governed by the underlying tokenizer. Tokenizers that process digits individually (such as Qwen's) yield a trie depth of at most $\lfloor\log_{10}K\rfloor+2$, while multi-digit tokenizers require depth two, enabling rapid parallel scoring across all branches (Appendix~\ref{app:tokenizers}).
\item \textbf{Preserving text generation.} The model retains the ability to emit structured answers in standard autoregressive text generation. We utilize this text mode to verify that the model adheres to the expected syntax post-training; fine-tuning leaves general generation capabilities intact (\S\ref{sec:scale}).
\end{itemize}

\subsection{Training-Free Recipe}
\label{sec:scoring}
\begin{figure}[t]
\centering
\includegraphics[width=\colfigwidth]{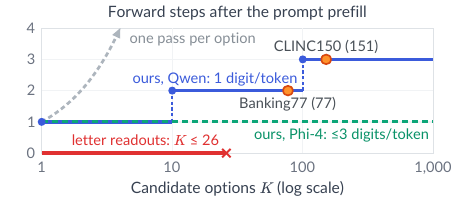}
\caption{Forward evaluation steps after prompt KV-cache prefill versus candidate count $K$.}
\label{fig:forward-steps}
\end{figure}

Let $x$ denote the input prompt ending with \texttt{Best answer: [}. We score candidate $k$ by accumulating suffix log-probabilities, followed by softmax normalization across all $K$ candidates:
\begin{equation}
\begin{aligned}
s_k&=\sum_{t=1}^{n_k}\log p_\theta(c_{k,t}\mid x,c_{k,<t}),\\
P(k\mid x)&=\frac{\exp(s_k)}{\sum_{j=1}^{K}\exp(s_j)}.
\end{aligned}
\label{eq:scoring}
\end{equation}

\paragraph{Joint contextual scoring.}
Conditioning suffix evaluations on the entire prompt allows the model to jointly evaluate all alternatives. In contrast, several community models score options in isolation~\citep{clm} or via separate binary prompts~\citep{yinsongxu2026llm2jev}. However, isolated evaluation struggles on queries requiring comparative judgment (such as finding the closest semantic match or ``none of the above''); on JevBench, isolated scoring trails joint readouts by about 5 points on identical backbones (76.2\% for LLM2Jev vs.\ 81.4\%; Table~\ref{tab:community}). While joint scoring introduces minor susceptibility to option order~\citep{zheng2024large}, we mitigate position bias during fine-tuning via random option shuffling.

\paragraph{Parallel suffix scoring.}
Because candidate suffixes are fixed and known in advance, they are scored concurrently via teacher forcing after caching the KV activations of $x$. As illustrated in Figure~\ref{fig:forward-steps}, even for 150+ candidates (e.g., CLINC150), evaluation requires only a single prefill and at most 1--3 brief forward steps without autoregressive decoding, drastically outperforming isolated $\mathcal{O}(K)$ scoring~\citep{clm,jevlocal}.

\subsection{Fine-Tuning Recipe}
\label{sec:objective}
\begin{figure*}[t]
\centering
\includegraphics[width=\textwidth]{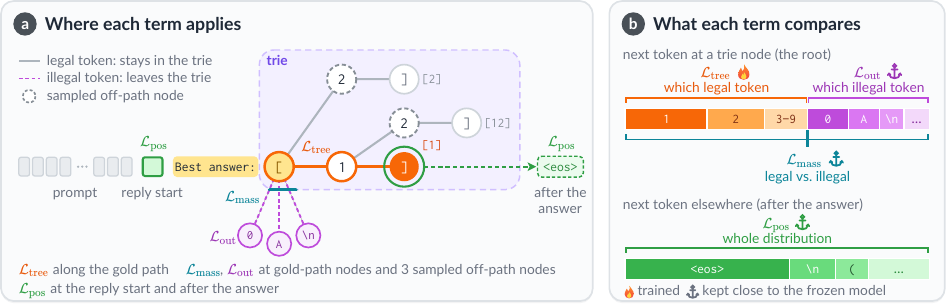}
\caption{Action sites of the tree-factorized listwise loss and KL regularization anchors.}
\label{fig:anchors}
\end{figure*}

\paragraph{Design objectives.}
Effective fine-tuning should optimize decision accuracy while preserving the base model's general conversational competence. To achieve minimal behavioral intervention, parameter updates should adjust only relative preferences among candidate options while leaving all other predictive behaviors intact. We achieve this by pairing a tree-factorized listwise loss with Kullback-Leibler (KL) divergence anchors against the base model $p_0$ (Figure~\ref{fig:anchors}).

\paragraph{Tree-factorized listwise loss.}
At each node $v$ of the candidate suffix trie, only a subset of vocabulary tokens form valid transitions. We designate these tokens as \emph{legal} ($\mathcal{L}(v)$) and define their aggregate probability as the \emph{legal mass}: $m_\theta(v)=\sum_{w\in\mathcal{L}(v)}p_\theta(w\mid x,v)$. Probabilities are locally renormalized over legal continuations along each branch:
\begin{equation}
q_\theta(k\mid x)=\prod_{t=1}^{n_k}\frac{p_\theta(c_{k,t}\mid x,c_{k,<t})}{m_\theta(c_{k,<t})}.
\label{eq:tree-prob}
\end{equation}
The decision loss is the negative log-likelihood of ground-truth candidate $y$: ${\mathcal{L}_{\text{tree}}=-\log q_\theta(y\mid x)}$. Crucially, evaluating $\mathcal{L}_{\text{tree}}$ requires computing probabilities solely along the \emph{gold path} (Figure~\ref{fig:anchors}a), making loss computation highly efficient even for high-cardinality tasks with hundreds of candidates. During inference, candidates are scored using the unnormalized joint likelihood in Eq.~\eqref{eq:scoring}; local renormalization in Eq.~\eqref{eq:tree-prob} is employed strictly during training to enable exact listwise cross-entropy optimization in a single forward pass.

\paragraph{KL divergence anchors.}
Because $\mathcal{L}_{\text{tree}}$ normalizes locally among legal tokens, it leaves illegal continuations and outside text positions unconstrained. Without regularization, the model's total legal mass can collapse or its non-decision distributions can drift arbitrarily. To safeguard behavioral fidelity, we introduce three complementary KL divergence anchors against the frozen reference model $p_0$ (Figure~\ref{fig:anchors}b):
\begin{equation}
\mathcal{L}_{\text{anchor}} = \mathcal{L}_{\text{mass}} + \mathcal{L}_{\text{out}} + \mathcal{L}_{\text{pos}},
\end{equation}
where $\mathcal{L}_{\text{mass}} = \sum_{v}\omega_v\,\mathrm{KL}_{\mathrm{B}}\big(m_0(v)\,\|\,m_\theta(v)\big)$ anchors legal vs.\ illegal mass at trie nodes; $\mathcal{L}_{\text{out}} = \sum_{v}\omega_v\,\mathrm{KL}\big(\bar p_0(\cdot\mid v)\,\|\,\bar p_\theta(\cdot\mid v)\big)$ preserves relative rankings among illegal exit tokens; and $\mathcal{L}_{\text{pos}} = \sum_{u}\omega_u\,\mathrm{KL}\big(p_0(\cdot\mid u)\,\|\,p_\theta(\cdot\mid u)\big)$ anchors full next-token distributions at context positions $u$. Weights $\omega_v$ allocate equal aggregate weight ($0.5$) to nodes along the gold path and to three randomly sampled off-path nodes; weights $\omega_u$ equally divide mass ($0.5$ each) between the reply start and the position immediately following the closing bracket. To save compute, we truncate anchored distributions to the top-64 tokens, capturing on average at least 94.5\% of probability mass with negligible memory overhead (Appendix~\ref{app:anchor-sites}).

\paragraph{Unified objective and training.}
The overall training objective balances candidate discrimination and base model preservation:
\begin{equation}
\mathcal{L}=\mathcal{L}_{\text{tree}}+\lambda\mathcal{L}_{\text{anchor}},
\end{equation}
where $\lambda$ governs anchor strength ($\lambda=1$ by default; ablations in \S\ref{sec:recipe}). We support both full-parameter tuning and parameter-efficient LoRA~\citep{hu2022lora} across all linear projections. Optimization uses AdamW with linear warmup and decay, candidate shuffling to counter position bias, and at most one training epoch (hyperparameters in Appendix~\ref{app:impl}).

\subsection{Training Data Construction}
\label{sec:data}
We curate training sets exclusively from publicly accessible, human-annotated datasets, converting each source into our unified decision template without altering ground-truth labels or soliciting synthetic labels from teacher models. To prevent train-test contamination, we filter all training instances that share a 13-word n-gram (or an identical sentence of six or more words) with any evaluation benchmark, and purge exact duplicates of validation inputs.

Training mixtures were developed iteratively: we began with canonical intent-classification tasks highlighted in early Jev demonstrations~\citep{typesafe2026jev}, subsequently incorporating reasoning and long-context corpora to expand coverage. Because subsequent data iterations were motivated by error analysis on public JevBench samples, we treat the evaluations in \S\ref{sec:reasoning-data} as exploratory; JevBench was strictly excluded from hyperparameter tuning and checkpoint selection. Appendix~\ref{app:finetune-details} details each data mixture, and Appendix~\ref{app:data} provides complete dataset breakdowns.

\section{Experiments}
\label{sec:experiments}

We outline the experimental setup in \S\ref{sec:setup}, followed by our core empirical findings: demonstrating that frozen LLMs are effective decision models out of the box (\S\ref{sec:training-free}), delineating the conditions under which fine-tuning provides tangible benefits (\S\ref{sec:finetuning}), and establishing optimal fine-tuning and regularization practices (\S\ref{sec:recipe}).

\subsection{Setup}
\label{sec:setup}

\paragraph{Foundation models.}
We conduct experiments on Qwen3.5-4B~\citep{qwen2026qwen35} (with thinking disabled except in \S\ref{sec:thinking}) and the compact Qwen3-0.6B~\citep{yang2025qwen3}. This choice aligns with prevalent community Jev-style models that build upon the same backbones~\citep{lee2026semif,plumb,hopper,reflex,jevk5,openalternativejev}, facilitating direct and fair comparisons. Furthermore, a dense 4B model incurs an inference cost of \$0.046/MTok (Table~\ref{tab:cost}, Appendix~\ref{app:cost}), closely matching Jev's official pricing (\$0.042/MTok) and establishing a realistic baseline for fast System~1 decision-making.

\paragraph{Training configurations.}
We evaluate six training configurations at 4B and five at 0.6B, spanning varying data mixtures and step budgets (\S\ref{sec:finetuning}; complete dataset compositions and step budgets in Tables~\ref{tab:datasets} and~\ref{tab:main}), all trained under the unified objective in \S\ref{sec:objective}. In \S\ref{sec:recipe}, we retrain \cfgLong{} across different anchor weights under both full-parameter tuning and LoRA.

\paragraph{Evaluation dimensions.}
We evaluate models across five complementary dimensions (full dataset specifications, splits, and citations in Appendix~\ref{app:eval} and Table~\ref{tab:eval-licenses}):
(1)~\textbf{External benchmarks}: seven classification and reasoning datasets retaining full candidate sets (BoolQ, MMLU, MMLU-Pro, ARC-C, WinoGrande, SciQ, and 77-way Banking77);
(2)~\textbf{JevBench}: 231 public items as an out-of-distribution diagnostic suite;
(3)~\textbf{General capabilities}: GSM8K, IFEval, TriviaQA, LAMBADA, and WikiText-2;
(4)~\textbf{Multimodal perception}: MMBench and MMStar; and
(5)~\textbf{Conversational behavior}: generation stability on Dolly-15k and MT-Bench.

\paragraph{Metrics and significance testing.}
We evaluate accuracy, ECE, and the Brier score (\S\ref{sec:related}). Model comparisons are validated using two-sided exact sign tests over the subset of instances on which predictions diverge.

\paragraph{Comparison baselines.}
We benchmark against four open-source community systems using their official implementations (SemIf, reflex 4B, open-alternative-jev, Winnow-12B) and proprietary GPT-5.6 Sol, reproducing official leaderboard results locally within four items.

\subsection{An LLM Is Already a Decision Model}
\label{sec:training-free}

\begin{table}[t]
\centering
\footnotesize
\setlength{\tabcolsep}{2.6pt}
\begin{tabular}{lrrrrr}
\toprule
 & \multicolumn{3}{c}{JevBench} & \multicolumn{2}{c}{External} \\
\cmidrule(lr){2-4}\cmidrule(lr){5-6}
System & Acc. & Hard & ECE & Macro & B77 \\
\midrule
GPT-5.6 Sol (generation) & 94.4 & 88.3 & -- & 91.5 & 86.0 \\
Winnow-12B$^\dagger$ & 86.6 & 74.8 & .058 & 82.2 & -- \\
\midrule
\textbf{4B training-free (ours)} & 81.4 & 64.9 & .057 & 78.6 & 69.0 \\
\textbf{4B fine-tuned (ours)}$^\dagger$ & 80.1 & 61.3 & .070 & 78.9 & 75.2 \\
\textbf{4B LoRA (ours)}$^\dagger$ & 84.0 & 68.5 & .050 & 78.9 & 75.4 \\
4B generation & 81.4 & 64.9 & -- & 78.4 & 68.5 \\
SemIf & 80.5 & 60.4 & .061 & 75.4 & -- \\
reflex 4B$^\dagger$ & 79.7 & 59.5 & .070 & 78.8 & -- \\
open-alternative-jev & 75.8 & 59.5 & .108 & 75.5 & -- \\
\midrule
\textbf{0.6B training-free (ours)} & 56.7 & 33.3 & .278 & 57.9 & 22.2 \\
\textbf{0.6B fine-tuned (ours)}$^\dagger$ & 58.0 & 37.8 & .118 & 59.1 & 60.5 \\
\textbf{0.6B LoRA (ours)}$^\dagger$ & 61.0 & 39.6 & .121 & 58.5 & 59.0 \\
0.6B generation & 56.7 & 33.3 & -- & 57.8 & 21.9 \\
\bottomrule
\end{tabular}
\caption{Accuracy and calibration across JevBench and External benchmarks (\%). $^\dagger$\,Fine-tuned; ours: \cfgLong{} (Table~\ref{tab:main}) or LoRA (${\lambda=1}$ at 4B, ${\lambda=0.01}$ at 0.6B; Table~\ref{tab:ablation}). Letter readouts cannot evaluate Banking77.}
\label{tab:training-free}
\end{table}

\paragraph{Strong zero-shot decision capability at 4B.}
Without task-specific adaptation, frozen Qwen3.5-4B achieves 81.4\% on public JevBench (188/231 correct; Table~\ref{tab:training-free}), on par with the training-free SemIf (80.5\%) and above the fine-tuned reflex 4B (79.7\%) on the same backbone. On External benchmarks, \sys{} outperforms SemIf by 3.2 percentage points in macro accuracy (313 wins vs.\ 154 losses), driven by solid gains on MMLU, MMLU-Pro, and ARC-C. Because both methods evaluate the exact same weights, these improvements validate the power of joint candidate conditioning and prefix-free scoring.

\paragraph{Effortless scaling to high-cardinality tasks.}
Many real-world tasks involve dozens or hundreds of candidates (e.g., Banking77 with 77 intents, CLINC150 with 151 classes). Here, letter-based readouts fail due to fixed 26-option limits, while isolated scoring models require dozens of separate forward passes. In contrast, \sys{} evaluates all candidates jointly in a single pass, achieving 69.0\% zero-shot accuracy on Banking77 with minimal overhead.

\paragraph{Calibrated probabilities without post-hoc scaling.}
The training-free 4B model achieves an ECE of 0.057 on JevBench out of the box, outperforming both SemIf (0.061) and reflex 4B (0.070, even with its post-hoc calibration mapping), confirming that raw next-token probabilities over numeric suffixes remain well-calibrated.

\paragraph{Multimodal visual decisions emerge zero-shot.}
Because our framework introduces no specialized heads or custom tokens, vision-language backbones evaluate multimodal queries out of the box through the identical interface. By simply prepending image tokens, Qwen3.5-4B attains 82.4\% on MMBench (CircularEval) and 64.6\% on MMStar without multimodal fine-tuning. Evaluating with visual inputs withheld drops performance to 15.2\% and 28.8\%, confirming genuine visual comprehension rather than reliance on language priors.

\begin{figure*}[t]
\centering
\includegraphics[width=\textwidth]{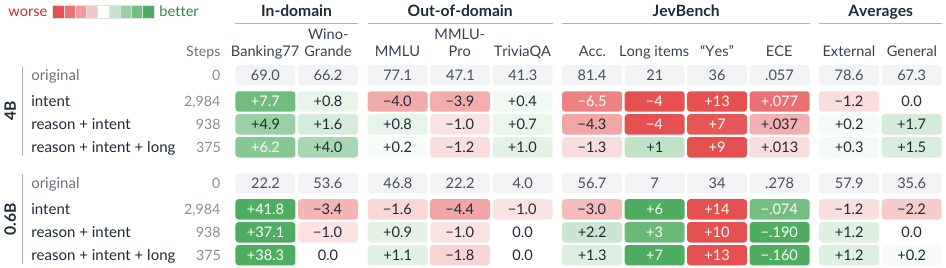}
\caption{Change from the original model after training on the intent (\cfgIL{}), reason + intent (\cfgRL{}), and reason + intent + long (\cfgLong{}) mixtures; color saturates at a change of 10 points. Table~\ref{tab:main} reports all configurations (six at 4B, five at 0.6B).}
\label{fig:finetune}
\end{figure*}

\subsection{When Fine-Tuning Helps}
\label{sec:finetuning}

To determine when parameter adaptation genuinely benefits decision models, we evaluate six training configurations at 4B and five at 0.6B (Table~\ref{tab:main}). As summarized in Figure~\ref{fig:finetune}, our empirical findings reveal three key principles:

\paragraph{Gains concentrate strictly on seen task domains.}
\label{sec:covered}
Performance improvements track the specific task structures covered during training, with little cross-domain transfer to unrepresented domains (Figure~\ref{fig:finetune}). Because every mixture incorporates intent classification, intent routing on Banking77 improves consistently across all configurations, gaining 4.9--7.7 points at 4B and 37--42 points at 0.6B. Conversely, knowledge-intensive benchmarks absent from the training mixtures, such as MMLU and TriviaQA, remain virtually flat regardless of training scale. Fine-tuning aligns output format and candidate discrimination, but cannot inject world knowledge that the pre-trained backbone lacks.

\paragraph{Format diversity and length alignment prevent heuristic shortcuts.}
\label{sec:intent-data}
\label{sec:reasoning-data}
Data composition and length diversity govern transfer fidelity far more than training volume alone:
\begin{itemize}
\item \emph{The Intent Trap}: Training solely on intent routing induces severe inductive bias: the model develops an aggressive confirmation bias toward ``yes'', with affirmative predictions surging on binary queries and suppressing fallback options, which degrades overall JevBench accuracy from 81.4\% to 74.9\%.
\item \emph{The Short-Context Trap}: Incorporating balanced reasoning corpora eliminates this affirmative bias and restores baseline accuracy. However, scaling short-context supervision alone impairs long-document comprehension, with correct predictions on long items dropping from 23 to 17.
\item \emph{Length Alignment}: Augmenting the mixture with long-context corpora halts this regression, restoring long-sequence accuracy at 4B and doubling it on 0.6B. Thus, effective transfer requires alignment in both task format and input length.
\end{itemize}

\paragraph{Model scale dictates the net return on adaptation.}
\label{sec:scale}
The benefits of fine-tuning depend fundamentally on backbone capacity:
\begin{itemize}
\item \emph{Compact models (0.6B)} experience broad, transformative gains: fine-tuning teaches basic format compliance, substantially boosting high-cardinality routing on Banking77 from 22.2\% to 60.5\% and improving calibration.
\item \emph{Capable models (4B)} already exhibit strong zero-shot decision boundaries and calibration. Consequently, full fine-tuning yields diminishing returns on broad out-of-distribution suites like JevBench, with gains restricted to targeted enterprise domains. Only parameter-efficient adaptation via LoRA improves overall accuracy (\S\ref{sec:lora}).
\end{itemize}
Importantly, fine-tuning at both scales preserves general linguistic capabilities and potentially other task capabilities (Appendix~\ref{app:finetune-details}), while 4B models fully retain zero-shot visual decisions on MMBench (Table~\ref{tab:image}).

\subsection{How to Fine-Tune}
\label{sec:recipe}

\begin{figure*}[t]
\centering
\includegraphics[width=\textwidth]{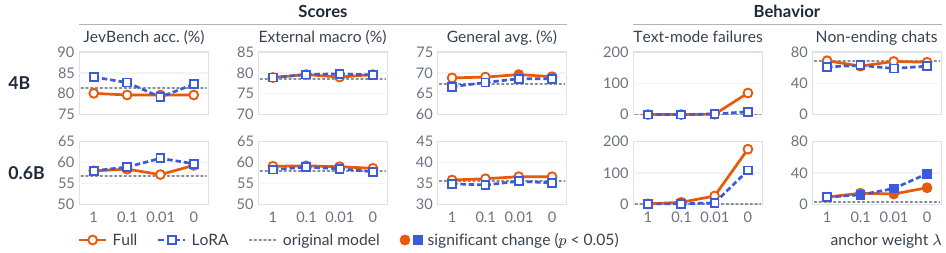}
\caption{Anchor-weight ($\lambda$) and LoRA ablations on \cfgLong{} across benchmark scores and generative behavior (Table~\ref{tab:ablation}).}
\label{fig:ablation}
\end{figure*}

Deploying decision tuning in practice involves two critical choices: selecting the regularization strength $\lambda$ and deciding between full-parameter tuning and parameter-efficient adapters. We evaluate these trade-offs by retraining \cfgLong{} across $\lambda \in \{1, 0.1, 0.01, 0\}$ under both regimes (Figure~\ref{fig:ablation}; Table~\ref{tab:ablation}).

\paragraph{Benchmark scores conceal generative collapse.}
Standard benchmark scores remain deceptively flat across anchor weights (Figure~\ref{fig:ablation}, left). However, omitting anchors (${\lambda=0}$) severely destabilizes autoregressive generation: unanchored 4B compulsively emits trailing text after closing brackets in 69 of 231 JevBench queries, while unanchored 0.6B suffers non-terminating chat loops (Figure~\ref{fig:ablation}, right). Setting ${\lambda=1}$ nearly eliminates format-breaking generations (none at 4B, one at 0.6B) while maintaining JevBench accuracy.

\paragraph{LoRA excels on capable backbones but requires anchors.}
\label{sec:lora}
For capable models, LoRA surpasses full-parameter tuning at three of the four anchor weights: 4B LoRA with ${\lambda=1}$ achieves 84.0\% on JevBench and the best calibration without thinking (ECE 0.050), mitigating representation drift~\citep{biderman2024lora}. On 0.6B, LoRA matches full tuning. Crucially, parameter efficiency cannot replace anchors: unanchored LoRA still suffers syntax and termination failures, confirming their complementary protective roles.

\subsection{Thinking Jev: Budgeted Test-Time Reasoning}
\label{sec:thinking}

\begin{figure}[t]
\centering
\includegraphics[width=\colfigwidth]{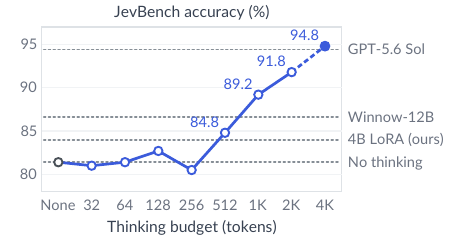}
\caption{Public JevBench accuracy of Qwen3.5-4B across thinking budgets with the \emph{Limit} prompt (Table~\ref{tab:thinking-full}). The 4,096-token point is a supplementary run on another GPU (Appendix~\ref{app:thinking}).}
\label{fig:thinking}
\end{figure}

While original Jev operates as an intuitive ``System~1'' model without intermediate reasoning, most remaining zero-shot errors concentrate on hard, multi-step queries where fine-tuning provides limited gains (\S\ref{sec:finetuning}). Because modern backbones natively support reasoning blocks, we introduce \textbf{Thinking Jev}: unlocking native test-time reasoning under a strict token budget $N$. We enforce the budget by cutting off thoughts exceeding $N$ tokens with \texttt{</think>}~\citep{muennighoff2025s1}, then append \texttt{Best answer: [} to extract candidate probabilities via Eq.~\eqref{eq:scoring}.

\paragraph{Accuracy scales with thinking budget.}
As shown in Figure~\ref{fig:thinking}, thoughts up to 256 tokens yield no significant gain because Qwen3.5-4B spends initial tokens on task restatement. Beyond 512 tokens, accuracy scales monotonically with budget, reaching 94.8\% on JevBench at 4,096 tokens (matching GPT-5.6 Sol at 94.4\%). These gains concentrate on hard multi-hop questions, and the final readout consistently selects the option reached by the thought.

\paragraph{Controlled latency and cost.}
While Thinking Jev introduces sequential decoding steps, its compute remains strictly budgeted: buying frontier accuracy on hard decisions while remaining over $5\times$ cheaper than large models in standard non-thinking mode (Appendix~\ref{app:cost}).

\section{Conclusion}
\label{sec:conclusion}

We presented \sys{}, an architecture-preserving framework turning off-the-shelf causal LLMs into calibrated Jev-style decision models through prefix-free conditioning, with or without parameter adaptation. Modern LLMs are inherently strong decision models: frozen backbones match or exceed existing training-free readouts of the same model, scale to arbitrary options, handle multimodal inputs, and unlock deliberate reasoning via Thinking Jev.

For practical deployment, we synthesize three core principles: (1)~\emph{Default to zero-shot}: foundation models provide robust baselines out of the box, with Thinking Jev scaling reasoning on demand; (2)~\emph{Aligned adaptation}: fine-tuning is most valuable for compact models or specialized routing when training distributions match test formats and lengths; and (3)~\emph{Anchor with LoRA}: pairing parameter-efficient LoRA with KL anchors (${\lambda=1}$) maximizes non-thinking 4B accuracy while preventing generative collapse. Together, these results establish a reliable, lightweight bridge between generative foundation models and deterministic decision systems.

\bibliographystyle{plainnat}
\bibliography{references}

\appendix
\section{Implementation Details}
\label{app:impl}

\paragraph{Prompt.}
The full user-message template is:
\begin{quote}\small\ttfamily
State: \{state\}\\
Options:\\
{}[1] \{option 1\}\\
{}[2] \{option 2\}\\
\dots\\
Select one option. Answer only with its bracketed numeric identifier.
\end{quote}
{\raggedright We apply the model's chat template with \texttt{add\_generation\_prompt=True} and \texttt{enable\_thinking=False}.\par}
\begin{itemize}
\item \textbf{Scoring mode} appends \texttt{Best answer: [} to the formatted prompt and computes the log-probability of each candidate suffix ``$k$\texttt{]}''.
\item \textbf{Text mode} continues the same prompt, including \texttt{Best answer: [}, with greedy generation for up to 32 new tokens and stops at EOS. The concatenated prefix and generated completion must begin with a valid candidate identifier \texttt{[$k$]}, which is extracted as the predicted answer; any trailing text is discarded, including output truncated at the length limit. Generations failing this leading syntax count as parse failures.
\end{itemize}
We take each candidate's suffix tokens from the tokenization of the full string, the prompt followed by the suffix, and check that the prompt's tokens form a prefix of it, so all candidates share one prompt prefill (Appendix~\ref{app:tokenizers}).

\paragraph{Reference statistics.}
Before training, we use the frozen original model to compute and cache the following anchor statistics for each example:
\begin{itemize}
\item the legal mass at each anchored node;
\item the top-64 illegal tokens and the mass of the remaining illegal tokens;
\item the top-64 tokens and the remaining mass at each anchored position.
\end{itemize}
Training reads these cached statistics, so it does not need a second model in memory. This top-$k$ truncation substantially compresses cache storage: retaining only the 64 most probable tokens and a single remainder bucket per anchored site requires negligible footprint, compared to storing full vocabulary distributions (151,936 tokens for Qwen3-0.6B; 248,320 for Qwen3.5-4B). Appendix~\ref{app:anchor-sites} provides systematic ablations over anchor sites, top-64 truncation, and off-path node sampling.

\paragraph{Latency.}
Our reference implementation uses FP32 and is not optimized for speed. On a single RTX A5500, median latency is 194\,ms for 4B and 76\,ms for 0.6B; SemIf takes 46\,ms in BF16. These measurements are indicative, and latency optimization is outside the scope of this work.

\paragraph{Hyperparameters.}
See Table~\ref{tab:hparams}.

\begin{table}[h]
\centering
\footnotesize
\setlength{\tabcolsep}{4pt}
\renewcommand{\arraystretch}{1.12}
\begin{tabular}{>{\raggedright\arraybackslash}p{2.2cm}>{\raggedright\arraybackslash}p{4.9cm}}
\toprule
Item & Setting \\
\midrule
Trainable parameters & language model + LM head (all) \\
LoRA (\S\ref{sec:lora}) & rank 16, $\alpha=32$, dropout 0, on decoder linear layers (32.5M trainable parameters for 4B, 10.1M for 0.6B); merged after training \\
Parallelism, precision & FSDP; BF16 compute, FP32 optimizer states \\
Optimizer & AdamW; weight decay 0.01; gradient clipping 1.0 \\
Peak learning rate & 4B: $2\times10^{-6}$; 0.6B: $5\times10^{-6}$; LoRA: $10^{-4}$ \\
Schedule & linear warmup (10\% of steps; 100 for one-epoch runs), linear decay to 10\% of the peak; one pass over the data \\
Global batch size & 32 (30 for \cfgIS) \\
Maximum length & 8,192 tokens; 10,240 for \cfgLong{} (inputs never truncated) \\
Anchors & 3 off-path nodes, top-64 \\
Loss weights & listwise 1; anchors ${\lambda=1}$ (\S\ref{sec:recipe}: 0.1, 0.01, 0) \\
Other & shuffled options; gradient checkpointing; seed 20260922 \\
Health check & stop if root legal log-mass drops $>0.5$ nats (ablations: warn only) \\
\bottomrule
\end{tabular}
\caption{Training hyperparameters.}
\label{tab:hparams}
\end{table}

\section{Evaluation Details}
\label{app:eval}

\paragraph{External.}
We use fixed samples totaling 7,080 items: BoolQ (500), MMLU (1,200), MMLU-Pro (800), ARC-Challenge (600), WinoGrande (500), SciQ (400), and the full Banking77 test set (3,080). We retain all options, including 10 for MMLU-Pro and 77 for Banking77. The External macro average is the mean accuracy on the first six benchmarks.

\paragraph{JevBench.}
We use the 231 public items, 37 of which are longer than 1,500 tokens. These items never appear in training, and we do not use them to tune learning rates, select checkpoints, or fit calibration.

\paragraph{General.}
We run lm-evaluation-harness~\citep{biderman2024lessons} on GSM8K (250 items, 5-shot, generated solutions), IFEval (200), TriviaQA (300, 5-shot, closed book), LAMBADA (500), and WikiText-2 (62 documents, perplexity). The General average is the mean accuracy on the first four tasks.

\paragraph{Image.}
Qwen3.5-4B also reads images, and training leaves its vision encoder unchanged. We place the image before the state and score the numbered options as usual. MMBench-EN v1.1 dev asks each of its 1,292 questions under every rotation of its options (4,876 prompts) and counts a question as correct only if all rotations are answered correctly (CircularEval). MMStar has 1,500 questions selected so that they require the image; we exclude two whose correct option is empty. We evaluate the original model and the largest configuration in each group of Table~\ref{tab:datasets}: \cfgIL{}, \cfgRL{}, and \cfgLong{}. Qwen3-0.6B has no vision encoder. We use VLMEvalKit's builds of both benchmarks~\citep{duan2024vlmevalkit}, the versions named in Qwen's model card, and verify them against VLMEvalKit's published checksums. The state holds a fixed instruction, MMBench's hint when present, and the question. All models use the original model's image processor. We downscale images to at most 1,048,576 pixels so that FP32 inference fits on 24\,GB GPUs; this affects 0.5\% of MMBench prompts and 2.7\% of MMStar prompts. Qwen3.5 assigns image tokens rotary positions from the image grid rather than from their order, so each candidate branch continues from the prompt's largest position; the cached scores match a full recomputation of every candidate to within $10^{-5}$. No 13-word window is shared between the image questions and any training set. Qwen reports 89.4 on MMBench and 78.3 on MMStar for generated answers~\citep{qwen2026qwen35}, so our scores are not directly comparable.

\paragraph{Behavior.}
We check the chat behavior of the ablation models (\S\ref{sec:recipe}). We take 50 prompts from each of Dolly-15k's eight categories in a fixed seeded order, dropping duplicates and prompts longer than 4,000 characters; any context follows the instruction after a blank line. Classification, closed QA, and information extraction form 150 short-answer prompts, where the right behavior is to answer and stop; the other five categories are long-form. MT-Bench adds 80 two-turn questions, whose second turn follows the model's own first reply, for 560 replies per model. Decoding uses the chat template with thinking disabled, BF16, greedy search, and no repetition penalty, and stops at the end-of-turn or end-of-text token or after 1,024 new tokens. Each original model and its fine-tuned models run on the same GPU type with the same batch size and order. We count replies that do not end within 1,024 tokens and repetition loops, in which at least half of the word 4-grams are repeats~\citep{welleck2020neural}. Some legitimate answers exceed 1,024 tokens (69 of the 4B original's replies), so we compare each reply with the original model's reply to the same prompt and count changes in both directions. The loop threshold flags 1 of the 4B original's replies and 6 of the 0.6B original's. Two Dolly prompts share a 13-word window with MuSiQue training paragraphs. Dolly-15k is released under CC BY-SA 3.0 and the MT-Bench prompts under Apache 2.0.

\paragraph{Metrics and compute.}
ECE uses 15 bins. Our models run in FP32 at inference time. Training runs use four A100 80GB GPUs, except \cfgIS{}, which uses three GPUs, a global batch size of 30, and 10 warmup steps. Training \cfgRS{} takes about 41 minutes of optimizer steps for 4B and 11 minutes for 0.6B.

\paragraph{Comparison systems.}
We run reflex 4B, a LoRA adapter, with its authors' calibration file, and Winnow-12B with 8-bit quantization (Q8).

\section{Tokenizer Differences}
\label{app:tokenizers}

\begin{figure*}[t]
\centering
\includegraphics[width=\textwidth]{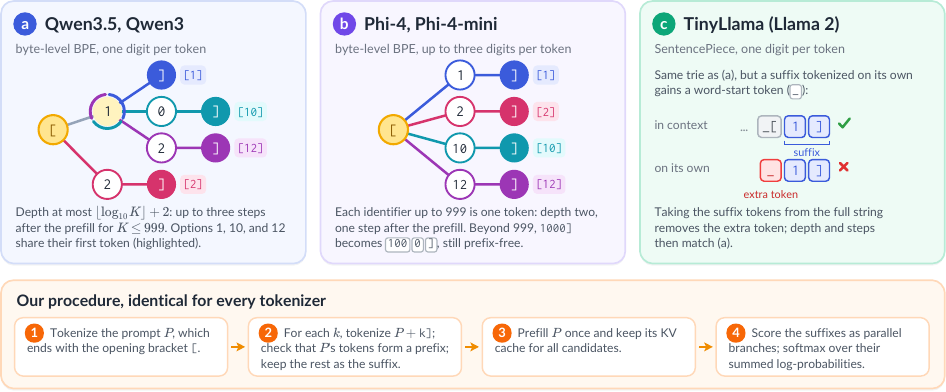}
\caption{How three tokenizer families split the candidate suffixes, and the procedure that handles all of them. (a) Digit-level tokenizers give a deeper trie in which options 1 and 10 share their first token. (b) Tokenizers that group up to three digits give a depth-two trie. (c) SentencePiece adds a word-start token when a suffix is tokenized on its own; taking suffix tokens from the full string (step 2) avoids it. We checked identifiers 1 to 1,000 with five tokenizers.}
\label{fig:tokenizers}
\end{figure*}

The \sys{} interface imposes only one formal invariant: the token sequence of the prompt prefix must strictly form a prefix of the full tokenization for each completed candidate. The structural topology of the candidate trie, however, depends on how the underlying tokenizer segments numeric strings. We systematically evaluated three widely used tokenizer families across a 1,000-option prompt (Figure~\ref{fig:tokenizers}); all three yield structurally prefix-free tries for identifiers 1 through 1,000.

\paragraph{Digit grouping.}
Qwen and Llama~2 tokenize digits individually, so an identifier with $d$ digits takes $d+1$ tokens, and the trie has depth $\lfloor\log_{10}K\rfloor+2$. Tokenizers that group up to three digits, such as those of Phi-4 and Phi-4-mini, make every identifier up to 999 a single token followed by the closing bracket, so the trie has depth two. Scoring then needs one short forward step after the prompt prefill instead of up to three, and the tree-factorized loss reduces to a flat cross-entropy over the legal first tokens. Beyond 999, identifiers span several tokens again (1000 becomes \tk{100}\tk{0}), but the closing bracket keeps the suffixes prefix-free, so the number of options remains unlimited.

\paragraph{Word-start markers.}
SentencePiece tokenizers such as that of Llama~2 treat the start of a string as the start of a word. Tokenized on its own, the suffix \texttt{1]} therefore gains a word-start token that it never has in context, where it follows the opening bracket. We take each suffix's tokens from the tokenization of the full string instead and check that the prompt's tokens form a prefix of it. The prompt tokens are then identical across candidates, so parallel decoding with one shared prompt prefill is unaffected. A tokenizer that merged the prompt's final token with the start of a suffix would fail this check; ending the prefill with an opening bracket avoids such merges for all tokenizers we checked. For Qwen, taking suffix tokens from the full string or from the suffix alone gives identical results.

\paragraph{Isolating the option-10 effect.}
With tokenizers that split digits individually, the suffixes of options 1 and 10 share their first token, and the next node chooses between the closing bracket and \texttt{0}. Tokenizers that group digits remove this shared node, so they offer a way to isolate the option-10 effect described in Appendix~\ref{app:option10}.

\section{Datasets}
\label{app:data}

Table~\ref{tab:datasets} lists the training datasets, their licenses, and example counts for each configuration, and Table~\ref{tab:eval-licenses} lists the licenses of the evaluation datasets. In addition to the overlap checks in \S\ref{sec:data}, we remove training examples whose inputs also appear in an official validation split.

\begin{table*}[t]
\centering
\footnotesize
\setlength{\tabcolsep}{3.1pt}
\begin{tabular}{l>{\raggedright\arraybackslash}p{2.75cm}rrrrrrl}
\toprule
Dataset & Task & Options & Intent & \twoline{Reason-}{9k} & \twoline{Reason-}{21k} & \twoline{Reason-}{30k} & \twoline{Reason+}{Long-12k} & License \\
\midrule
CLINC150 & intent (10 domains) & 151 & 15,207 & 1k & 1k & 1k & 1k & CC BY 3.0 \\
MASSIVE (en-US) & intent (voice assistant) & 60 & 11,390 & 1k & 1k & 1k & 1k & CC BY 4.0 \\
Bitext & intent (e-commerce support) & 27 & 21,839 & 1k & 1k & 1k & 1k & CDLA-Sharing-1.0 \\
CommonsenseQA & commonsense QA & 5 & 9,139 & -- & -- & -- & -- & MIT \\
HellaSwag & situation completion & 4 & 37,886 & -- & -- & -- & -- & MIT \\
\midrule
ReClor & argument reasoning & 4 & -- & 1k & 3k & 3k & 1k & research only \\
LogiQA 2.0 & logical reasoning & 4 & -- & 1k & 3k & 3k & 1k & CC BY-NC-SA 4.0 \\
CosmosQA & commonsense reading & 4 & -- & 1k & 3k & 3k & 1k & CC BY 4.0 \\
PIQA & physical commonsense & 2 & -- & 1k & 3k & 3k & 1k & AFL-3.0 \\
$\alpha$NLI & abductive reasoning & 2 & -- & 1k & 3k & 3k & 1k & not stated \\
ProofWriter & rule verification & 3 & -- & 1k & 3k & 3k & 1k & not stated \\
\midrule
VitaminC & fact verification & 3 & -- & -- & -- & 3k & -- & CC BY-SA 3.0 \\
QuAIL & reading comprehension & 4 & -- & -- & -- & 3k & -- & CC BY-NC-SA 4.0 \\
Social IQa & social commonsense & 3 & -- & -- & -- & 3k & -- & CC BY 4.0 \\
\midrule
ContractNLI & contract inference (long) & 3 & -- & -- & -- & -- & 1k & CC BY 4.0 \\
MuSiQue & multi-hop answerability (long) & 2 & -- & -- & -- & -- & 1k & CC BY 4.0 \\
QuALITY & long-document reading & 4 & -- & -- & -- & -- & 1k & CC BY 4.0 \\
\midrule
Total & & & 95,461 & 9,000 & 21,000 & 30,000 & 12,000 & \\
\bottomrule
\end{tabular}
\caption{Datasets per training configuration (citations in Appendix~\ref{app:finetune-details}). \cfgIL{} trains on the full intent set and \cfgIS{} (4B only) on 3,000 of its examples. Median input lengths for long-context datasets are 2.1k (ContractNLI), 2.4k (MuSiQue), and 6.8k tokens (QuALITY). ReClor permits non-commercial research use only; license declarations were absent from official $\alpha$NLI and ProofWriter releases.}
\label{tab:datasets}
\end{table*}

\begin{table*}[t]
\centering
\footnotesize
\begin{tabular}{llll}
\toprule
Suite & Dataset & Reference & License \\
\midrule
JevBench & 231 public items & \citet{benchmarkheaven2026jevbench} & MIT \\
\midrule
\multirow{7}{*}{External} & BoolQ & \citet{clark2019boolq} & CC BY-SA 3.0 \\
 & MMLU & \citet{hendrycks2021measuring} & MIT \\
 & MMLU-Pro & \citet{wang2024mmlupro} & MIT \\
 & ARC-Challenge & \citet{clark2018think} & CC BY-SA 4.0 \\
 & WinoGrande & \citet{sakaguchi2020winogrande} & CC BY \\
 & SciQ & \citet{welbl2017crowdsourcing} & CC BY-NC 3.0 \\
 & Banking77 & \citet{casanueva2020efficient} & CC BY 4.0 \\
\midrule
\multirow{5}{*}{General} & GSM8K & \citet{cobbe2021training} & MIT \\
 & IFEval & \citet{zhou2023instruction} & Apache 2.0 \\
 & TriviaQA & \citet{joshi2017triviaqa} & Apache 2.0 \\
 & LAMBADA & \citet{paperno2016lambada} & CC BY 4.0 \\
 & WikiText-2 & \citet{merity2017pointer} & CC BY-SA 3.0 \\
\midrule
\multirow{2}{*}{Image} & MMBench & \citet{liu2024mmbench} & Apache 2.0 \\
 & MMStar & \citet{chen2024mmstar} & not stated \\
\midrule
\multirow{2}{*}{Behavior} & Dolly-15k & \citet{conover2023free} & CC BY-SA 3.0 \\
 & MT-Bench & \citet{zheng2023judging} & Apache 2.0 \\
\bottomrule
\end{tabular}
\caption{References and licenses for official evaluation benchmark releases. MMStar does not provide an explicit license statement.}
\label{tab:eval-licenses}
\end{table*}

\begin{table}[h]
\centering
\footnotesize
\setlength{\tabcolsep}{3pt}
\begin{tabular}{@{}lrrr@{}}
\toprule
 & Intent & Reasoning & JevBench \\
\midrule
Examples & 95,461 & 9,000 & 231 \\
Intent classification & 51\% & 33\% & 10\% \\
Two-option & 0\% & 22\% & 32\% \\
Verification & 0\% & 11\% & 32\% \\
Answer ``none''/``unknown'' & 0.3\% & 6.0\% & 3.5\% \\
\bottomrule
\end{tabular}
\caption{Composition of the intent set, the reasoning set (\cfgRS{}), and the 231 public JevBench items. Verification includes yes/no and true/false/unknown questions.}
\label{tab:mix}
\end{table}

\section{Similarity Between Training and Evaluation Data}
\label{app:similarity}

\paragraph{Data contamination checks.}
We enforce strict separation between training corpora and evaluation suites by systematically checking for verbatim overlap:
\begin{itemize}
\item \textbf{Reasoning data vs.\ External and JevBench.} No 13-word window is shared between the 9,000 training examples and the 7,311 External and JevBench items. Before sampling, the full training sets contained one shared passage: an argument in LogiQA 2.0 also appears in an MMLU item and its MMLU-Pro counterpart. Our overlap filter removed it from the sample.
\item \textbf{Intent data.} The intent datasets share some label names with Banking77 and JevBench intent items (e.g., \texttt{cancel\_order}), but no user utterances.
\item \textbf{General suite.} The same check finds no overlap for the intent set, \cfgRS{}, \cfgRM{}, or \cfgRL{}. \cfgLong{} has two incidental overlaps from MuSiQue's Wikipedia paragraphs: one with an IFEval prompt, where evaluation concerns only the output format, and one with a WikiText article that shares a paragraph with a MuSiQue example.
\end{itemize}

\paragraph{Lexical similarity.}
After removing instruction templates, we compute TF-IDF vectors over content words. We weight each dataset by its share of the training mixture and compute cosine similarity to each evaluation set (Table~\ref{tab:similarity}).
\begin{itemize}
\item The reasoning set has higher lexical similarity to all External reasoning benchmarks, largely because ReClor and LogiQA resemble MMLU.
\item The intent set has higher similarity to Banking77 and JevBench.
\end{itemize}
Together with the results in \S\ref{sec:covered}, these comparisons suggest that shared task formats explain the observed gains better than shared vocabulary.

\begin{table}[h]
\centering
\footnotesize
\begin{tabular}{lrr}
\toprule
Evaluation set & Intent & Reasoning \\
\midrule
Banking77 & \textbf{.553} & .444 \\
BoolQ & .243 & \textbf{.332} \\
MMLU & .241 & \textbf{.337} \\
MMLU-Pro & .256 & \textbf{.322} \\
ARC-Challenge & .227 & \textbf{.264} \\
WinoGrande & .098 & \textbf{.132} \\
SciQ & .189 & \textbf{.222} \\
JevBench (public items) & \textbf{.295} & .258 \\
\bottomrule
\end{tabular}
\caption{TF-IDF cosine similarity between evaluation benchmarks and training mixtures.}
\label{tab:similarity}
\end{table}

\section{Full Results}
\label{app:results}

Table~\ref{tab:main} gives the fine-tuning results summarized in Figure~\ref{fig:finetune}. Tables~\ref{tab:eval1-full}, \ref{tab:eval3-full}, and~\ref{tab:image} report per-task results for the External, General, and Image suites.

\begin{table*}[t]
\centering
\footnotesize
\setlength{\tabcolsep}{3.9pt}
\begin{tabular}{llrrrrrrrrrr}
\toprule
 & & & \multicolumn{6}{c}{JevBench (231 public items)} & \multicolumn{2}{c}{External} & General \\
\cmidrule(lr){4-9}\cmidrule(lr){10-11}\cmidrule(lr){12-12}
Model & Config & Steps & Acc. & Won/lost ($p$) & Hard & Long/37 & ECE & ``Yes'' & Macro & B77 & Avg. \\
\midrule
\multirow{7}{*}{4B}
 & original & 0 & \textbf{81.4} & -- & \textbf{72} & 21 & \textbf{.057} & 36 & 78.6 & 69.0 & 67.3 \\
 & \cfgIS & 100 & 77.5 & 12/21 (.16) & 65 & 22 & .092 & 55 & 78.3 & 74.6 & 68.2 \\
 & \cfgIL & 2,984 & 74.9 & 8/23 (.01) & 60 & 17 & .134 & 49 & 77.4 & \textbf{76.7} & 67.3 \\
 & \cfgRS & 282 & \textbf{81.4} & 14/14 (1.0) & 70 & \textbf{23} & .067 & 44 & 78.7 & 75.9 & 68.8 \\
 & \cfgRM & 657 & 78.8 & 11/17 (.34) & 64 & 19 & .094 & 43 & \textbf{79.3} & 74.8 & 67.8 \\
 & \cfgRL & 938 & 77.1 & 10/20 (.10) & 60 & 17 & .094 & 43 & 78.8 & 74.0 & \textbf{69.0} \\
 & \cfgLong & 375 & 80.1 & 13/16 (.71) & 68 & 22 & .070 & 45 & 78.9 & 75.2 & 68.8 \\
\midrule
\multirow{6}{*}{0.6B}
 & original & 0 & 56.7 & -- & 37 & 7 & .278 & 34 & 57.9 & 22.2 & 35.6 \\
 & \cfgIL & 2,984 & 53.7 & 22/29 (.40) & 41 & 13 & .204 & 48 & 56.7 & \textbf{64.0} & 33.4 \\
 & \cfgRS & 282 & 55.0 & 11/15 (.56) & 36 & 9 & .118 & 41 & \textbf{59.1} & 60.4 & \textbf{36.8} \\
 & \cfgRM & 657 & 58.0 & 17/14 (.72) & 40 & 11 & .104 & 42 & 58.6 & 60.1 & 35.6 \\
 & \cfgRL & 938 & \textbf{58.9} & 17/12 (.46) & 41 & 10 & \textbf{.088} & 44 & 59.0 & 59.4 & 35.6 \\
 & \cfgLong & 375 & 58.0 & 17/14 (.72) & \textbf{42} & \textbf{14} & .118 & 47 & \textbf{59.1} & 60.5 & 35.8 \\
\bottomrule
\end{tabular}
\caption{Fine-tuning performance across training configurations (accuracy in \%; Hard and Long denote correct counts on JevBench subsets, with Long covering 37 items $>1,500$ tokens). Won/lost pairs report changes relative to base models with two-sided sign-test $p$-values. ``Yes'' counts affirmative predictions on 74 binary items (35 ground-truth ``yes''). General averages four core language benchmarks. In pooled testing, only 4B \cfgRL{} shows a significant shift on General (${p=0.03}$).}
\label{tab:main}
\end{table*}

\begin{table*}[t]
\centering
\footnotesize
\setlength{\tabcolsep}{3.6pt}
\begin{tabular}{llrrrrrrrrr}
\toprule
 & System / config & BoolQ & MMLU & Pro & ARC-C & Wino. & SciQ & Macro & B77 & ECE \\
\midrule
\multirow{5}{*}{Reference} & GPT-5.6 Sol (generation) & 91.80 & 91.42 & 77.62 & 98.17 & 90.40 & 99.50 & 91.48 & 86.04 & -- \\
 & Winnow-12B (Q8) & 89.20 & 79.67 & 56.75 & 95.17 & 72.60 & 100.00 & 82.23 & -- & .080 \\
 & reflex 4B (LoRA) & 89.00 & 75.00 & 52.50 & 92.00 & 64.80 & 99.50 & 78.80 & -- & .056 \\
 & SemIf & 88.00 & 71.50 & 40.25 & 90.00 & 62.80 & 99.75 & 75.38 & -- & .080 \\
 & open-alternative-jev & 87.80 & 71.92 & 41.50 & 89.50 & 63.00 & 99.50 & 75.54 & -- & .096 \\
\midrule
\multirow{8}{*}{4B} & original (training-free) & 88.00 & 77.08 & 47.12 & 93.33 & 66.20 & 99.75 & 78.58 & 69.03 & .058 \\
 & original (generation) & 88.00 & 77.08 & 46.00 & 93.33 & 66.20 & 99.75 & 78.39 & 68.54 & -- \\
 & \cfgIS & 89.00 & 75.92 & 44.25 & 92.67 & 67.80 & 100.00 & 78.27 & 74.64 & .085 \\
 & \cfgIL & 88.60 & 73.08 & 43.25 & 92.17 & 67.00 & 100.00 & 77.35 & 76.72 & .101 \\
 & \cfgRS & 88.60 & 76.92 & 45.50 & 92.50 & 68.80 & 99.75 & 78.68 & 75.91 & .067 \\
 & \cfgRM & 88.60 & 76.67 & 48.00 & 93.17 & 69.60 & 100.00 & 79.34 & 74.84 & .089 \\
 & \cfgRL & 88.40 & 77.92 & 46.12 & 92.67 & 67.80 & 100.00 & 78.82 & 73.96 & .076 \\
 & \cfgLong & 88.00 & 77.25 & 45.88 & 92.00 & 70.20 & 100.00 & 78.89 & 75.19 & .087 \\
\midrule
\multirow{7}{*}{0.6B} & original (training-free) & 72.60 & 46.75 & 22.25 & 54.17 & 53.60 & 97.75 & 57.85 & 22.21 & .272 \\
 & original (generation) & 72.60 & 46.67 & 22.12 & 54.17 & 53.60 & 97.75 & 57.82 & 21.95 & -- \\
 & \cfgIL & 73.80 & 45.17 & 17.88 & 57.17 & 50.20 & 95.75 & 56.66 & 63.99 & .158 \\
 & \cfgRS & 75.20 & 48.17 & 20.12 & 59.17 & 53.80 & 98.25 & 59.12 & 60.42 & .082 \\
 & \cfgRM & 74.60 & 48.17 & 21.38 & 59.17 & 50.00 & 98.00 & 58.55 & 60.10 & .106 \\
 & \cfgRL & 72.00 & 47.67 & 21.25 & 62.00 & 52.60 & 98.75 & 59.04 & 59.35 & .097 \\
 & \cfgLong & 74.40 & 47.83 & 20.50 & 60.33 & 53.60 & 97.75 & 59.07 & 60.49 & .100 \\
\bottomrule
\end{tabular}
\caption{Detailed accuracy on External benchmarks (\%). Macro averages the first six tasks; ECE reports 15-bin calibration across supported tasks. Pro: MMLU-Pro; Wino: WinoGrande; B77: Banking77.}
\label{tab:eval1-full}
\end{table*}

\begin{table*}[t]
\centering
\footnotesize
\setlength{\tabcolsep}{4pt}
\begin{tabular}{llrrrrrrrr}
\toprule
 & Config & Steps & Avg. & GSM8K & IFEval & TriviaQA & LAMBADA & LAMBADA ppl & WikiText ppl \\
\midrule
\multirow{7}{*}{4B} & original & 0 & 67.3 & 84.0 & 80.5 & 41.3 & 63.2 & 5.11 & 11.47 \\
 & \cfgIS & 100 & 68.2 & 85.6 & 82.0 & 42.0 & 63.4 & 5.11 & 11.44 \\
 & \cfgIL & 2,984 & 67.3 & 84.0 & 79.0 & 41.7 & 64.6 & 4.78 & 11.62 \\
 & \cfgRS & 282 & 68.8 & 86.0 & 84.0 & 42.0 & 63.0 & 5.06 & 11.42 \\
 & \cfgRM & 657 & 67.8 & 84.4 & 82.0 & 40.7 & 64.0 & 5.00 & 11.43 \\
 & \cfgRL & 938 & 69.0$^*$ & 86.4 & 82.5 & 42.0 & 65.0 & 5.03 & 11.47 \\
 & \cfgLong & 375 & 68.8 & 87.2 & 82.0 & 42.3 & 63.6 & 4.99 & 11.44 \\
\midrule
\multirow{6}{*}{0.6B} & original & 0 & 35.6 & 44.4 & 55.0 & 4.0 & 39.0 & 28.21 & 27.52 \\
 & \cfgIL & 2,984 & 33.4 & 43.2 & 48.0$^*$ & 3.0 & 39.2 & 24.34 & 26.85 \\
 & \cfgRS & 282 & 36.8 & 48.4 & 54.0 & 4.0 & 40.8 & 25.14 & 26.31 \\
 & \cfgRM & 657 & 35.6 & 44.8 & 54.5 & 2.7 & 40.4 & 25.80 & 26.37 \\
 & \cfgRL & 938 & 35.6 & 44.0 & 54.0 & 4.0 & 40.4 & 27.74 & 27.01 \\
 & \cfgLong & 375 & 35.8 & 48.4 & 51.0 & 4.0 & 39.8 & 26.50 & 26.64 \\
\bottomrule
\end{tabular}
\caption{General language modeling and reasoning capabilities (accuracy in \%; WikiText perplexity). Avg.\ denotes four-task mean accuracy. $^*$\,Statistically significant shift from the base model (${p<0.05}$).}
\label{tab:eval3-full}
\end{table*}

\paragraph{Image results.}
Table~\ref{tab:image} gives the results. None of the three fine-tuned models suffers performance degradation on either benchmark, with all three configurations significantly improving MMBench accuracy: \cfgIL{} reaches 84.1\% (57 wins vs.\ 35 losses; ${p=0.03}$), \cfgRL{} reaches 85.3\% (57 vs.\ 19; ${p<0.001}$), and \cfgLong{} reaches 84.9\% (49 vs.\ 16; ${p<0.001}$). On MMStar, accuracy increases modestly by 0.9--1.7 points (non-significant across all runs). Almost all MMBench questions that fine-tuning gains are ones the original model answers correctly under some rotations but not others: 48 of 57 for \cfgIL{}, 52 of 57 for \cfgRL{}, and 48 of 49 for \cfgLong{}. The fine-tuned models also choose the same option under every rotation more often (90.2--91.3\% of questions vs.\ 87.6\%), consistent with training on shuffled options. Each benchmark has six categories. Coarse perception on MMBench improves significantly for all three models; of the other 33 model--category pairs, two change significantly, about as many as chance would produce: MMBench relation reasoning rises for both \cfgRL{} and \cfgLong{} (8 wins, 1 loss each). Calibration exhibits minor variance: MMStar ECE ranges from 0.053 (\cfgRL{}) to 0.093 (\cfgIL{}) relative to 0.090 for the base model, and MMBench ECE stays between 0.014 and 0.019 (0.014 for the base model).

\begin{table*}[t]
\centering
\footnotesize
\setlength{\tabcolsep}{4pt}
\begin{tabular}{lrrrrrrrr}
\toprule
 & \multicolumn{5}{c}{MMBench-EN v1.1 dev (1,292 questions)} & \multicolumn{3}{c}{MMStar (1,498 questions)} \\
\cmidrule(lr){2-6}\cmidrule(lr){7-9}
Model & Acc. & Won/lost & Coarse & Same & ECE & Acc. & Won/lost & ECE \\
\midrule
original & 82.4 & -- & 81.2 & 87.6 & .014 & 64.6 & -- & .090 \\
\cfgIL & 84.1$^*$ & 57/35 & 86.5$^*$ & 90.5 & .014 & 65.5 & 126/112 & .093 \\
\cfgRL & 85.3$^*$ & 57/19 & 85.6$^*$ & 91.3 & .018 & 65.9 & 113/93 & .053 \\
\cfgLong & 84.9$^*$ & 49/16 & 84.5$^*$ & 90.2 & .019 & 66.3 & 108/82 & .086 \\
\midrule
original, no image & 15.2 & -- & 9.1 & 30.8 & .214 & 28.8 & -- & .234 \\
uniform guess & 2.2 & -- & 2.5 & -- & -- & 26.9 & -- & -- \\
\bottomrule
\end{tabular}
\caption{Zero-shot visual decision performance for Qwen3.5-4B (accuracy in \%). MMBench reports CircularEval accuracy (requiring correct predictions across all option permutations). Coarse denotes the 362 coarse-perception queries; Same measures prediction invariance across rotations. $^*$\,Statistically significant shift from the base model (${p<0.05}$).}
\label{tab:image}
\end{table*}

\section{Details of the Data Comparisons}
\label{app:finetune-details}

This appendix provides extended qualitative and empirical analyses supporting the findings in \S\ref{sec:finetuning}.

\paragraph{Intent data curation.}
To leverage high-quality human supervision, we began with three public intent-classification benchmarks featuring extensive label sets: CLINC150~\citep{larson2019evaluation} (151 intents, including an explicit out-of-scope category), MASSIVE en-US~\citep{fitzgerald2023massive} (60 intents), and Bitext customer support~\citep{bitext2024customer} (27 intents). We added CommonsenseQA~\citep{talmor2019commonsenseqa} and HellaSwag~\citep{zellers2019hellaswag} to cover general commonsense reasoning and narrative continuation. Following cleaning and deduplication, this \emph{intent set} comprises 95,461 instances, roughly half of which represent multi-class intent queries.

\paragraph{Impact of training duration.}
We compared 100 steps on the intent set (\cfgIS{}, 3,000 examples) with one full epoch (\cfgIL{}, 2,984 steps). For 4B, longer training makes transfer worse: JevBench accuracy falls from 81.4\% to 77.5\% after 100 steps and to 74.9\% after the full epoch (8 wins, 23 losses, ${p=0.01}$). The full-epoch run also reduces External macro accuracy by 1.2 points (pooled ${p<0.001}$). Yet the in-distribution development loss continues to decrease monotonically (Bitext: 0.65 $\rightarrow$ 0.001), while the development check based on text generation plateaus at about 500 steps.

\paragraph{Systematic error patterns.}
Qualitative error analysis of individual predictions from \cfgIL{} reveals four systematic failure modes:
\begin{itemize}
\item \textbf{A bias toward ``yes''.} The probability of ``yes'' increases on 58 of the 74 yes/no items. The number of ``yes'' predictions rises from 36 to 49, although only 35 items have ``yes'' as the correct answer. Of the 11 new yes/no errors, 10 occur when the correct answer is ``no''.
\item \textbf{Avoiding ``none of these''.} The mean probability assigned to catch-all options falls from 0.056 to 0.010.
\item \textbf{Keyword matching.} On hard multiple-choice items, the model more often follows misleading keyword cues in the question.
\item \textbf{Overconfidence.} ECE rises from 0.057 to 0.134, and mean confidence in incorrect answers rises from 0.60 to 0.73.
\end{itemize}
Meanwhile, the same runs gain 5.6 and 7.7 points on Banking77: the data teaches what it covers, and the failures come from what it leaves out. The training data offers plausible explanations for these patterns. It contains no yes/no or verification questions, most intent examples can be solved by keyword matching, and all examples use just four fixed instruction templates. ``None of these'' is correct in only 250 of the 95,461 examples. Even \cfgIS{} shows a ``yes'' bias (55 ``yes'' answers), pointing to data composition rather than training duration alone. The same intent data produces a similar bias at 0.6B (48 ``yes'' answers).

\paragraph{Reasoning data.}
To counter these empirical failure modes, we curate a diverse mixture of six reasoning datasets (R6). ProofWriter~\citep{tafjord2021proofwriter} supplies rule-based verification with balanced true, false, and unknown answers, filling a gap in the intent set and countering the ``yes'' bias. CosmosQA~\citep{huang2019cosmos} adds questions where ``none of the above'' is correct; we upsample these to 16\%. ReClor~\citep{yu2020reclor} and LogiQA 2.0~\citep{liu2023logiqa2} require reasoning about arguments rather than matching keywords. PIQA~\citep{bisk2020piqa} and $\alpha$NLI~\citep{bhagavatula2020abductive} add two-option commonsense questions. Two-option questions account for 32\% of JevBench but were absent from the intent set. We retain the three intent-classification datasets (I3) to preserve intent skills and remove CommonsenseQA and HellaSwag. Because the intent-data experiments showed early in-domain gains, we start with only 1,000 examples per dataset and a single pass (\cfgRS{}, 282 steps). Table~\ref{tab:mix} compares the two mixtures.

\paragraph{Impact of format-balanced data.}
This mixture successfully curbs the affirmative bias, improves calibration relative to intent tuning, and matches zero-shot JevBench accuracy: with \cfgRS{}, the 4B model achieves 81.4\% (14 wins, 14 losses), marking the top result among all full-parameter 4B configurations. Its ECE of 0.067 represents the best calibration among full-parameter 4B runs (though slightly trailing the base model's 0.057), while ``yes'' predictions drop to 44. External macro accuracy reaches 78.7\% (+0.1 points). Among individual tasks, only Banking77 shifts significantly (+6.9 points, ${p<10^{-20}}$), while variations on WinoGrande (+2.6) and MMLU-Pro ($-1.6$) remain within statistical margins.

\paragraph{Scaling short-context reasoning data.}
We expand the reasoning set in two ways. \cfgRM{} uses 3,000 examples per reasoning dataset (657 steps). \cfgRL{} adds three more datasets with 3,000 examples each (938 steps): VitaminC~\citep{schuster2021get} for fact verification, QuAIL~\citep{rogers2020getting} for reading comprehension with ``not enough information'' answers, and Social IQa~\citep{sap2019social} for social commonsense. Neither improves the 4B model's JevBench accuracy, which falls to 78.8\% and 77.1\%, respectively. Against \cfgRS{}, \cfgRM{} has 0 wins and 6 losses (${p=0.03}$) and \cfgRL{} 0 wins and 10 losses (${p=0.002}$). At 0.6B, more reasoning data does not hurt: \cfgRL{} wins 14 items and loses 5 against \cfgRS{} (${p=0.06}$). External macro accuracy remains nearly unchanged (78.7--79.3\%).

\paragraph{Vulnerability on long-context queries.}
The losses concentrate on the 37 long items. \cfgRS{} answers 23 correctly, compared with 19 for \cfgRM{} (${p=0.13}$) and 17 for \cfgRL{} (${p=0.03}$); 6 of the 10 items that \cfgRL{} loses against \cfgRS{} are long, although long items make up only 16\% of JevBench. Both the intent and reasoning mixtures contain prompts no longer than about 570 words, whereas hard JevBench items include long policy documents and multi-hop questions. This mismatch suggests that more training on short inputs can hurt performance on long ones.

\paragraph{Long-input data.}
To address the length mismatch, we add three long-input reasoning datasets to \cfgRS{}, with 1,000 examples each: ContractNLI~\citep{koreeda2021contractnli} for entailment over full contracts, MuSiQue~\citep{trivedi2022musique} for deciding whether 20 paragraphs contain enough information to answer a multi-hop question, and QuALITY~\citep{pang2022quality} for questions about long articles. Their median input lengths range from 2.1k to 6.8k tokens. The resulting configuration, \cfgLong{}, uses 12,000 examples and 375 steps.

\paragraph{Restoration via long-sequence supervision.}
At the 4B scale, incorporating long-context data restores long-sequence accuracy to baseline levels. Specifically, \cfgLong{} correctly answers 22 long items, comparable to 21 for the base model and 23 for \cfgRS{} (yielding 6 wins against 1 loss relative to \cfgRL{}; ${p=0.13}$), and attains 80.1\% on JevBench (9 wins and 2 losses relative to \cfgRL{}; ${p=0.065}$). While it does not surpass \cfgRS{} (81.4\%; 1 win, 4 losses, ${p=0.38}$), which remains our strongest full-parameter 4B configuration, long-context supervision successfully reverses degradation without over-fitting the original model or \cfgRS{}. \cfgLong{} also trains for fewer steps than \cfgRL{} (375 vs.\ 938), which may contribute to the recovery. At 0.6B, long-context supervision provides decisive gains: \cfgLong{} resolves 14 long items correctly, doubling the base model's score (7 wins, 0 losses; ${p=0.02}$) and outperforming all other training configurations (9--13).

\paragraph{The 0.6B model gains broadly.}
Banking77 accuracy rises by 37.1--41.8 points, \cfgRS{} reduces CLINC150 development loss from 6.91 to 0.91, and ARC-Challenge gains 3.0--7.8 points. Calibration improves as well: JevBench ECE falls from 0.278 to 0.088--0.204, and mean External ECE from 0.272 to 0.082--0.158. JevBench accuracy stays close to the original (53.7--58.9\% vs.\ 56.7\%; best: \cfgRL{}), and the best 0.6B result is LoRA at ${\lambda=0.01}$ (61.0\%; \S\ref{sec:lora}).

\paragraph{The 4B model gains only where it is weak.}
No full-parameter configuration beats the original 4B model on JevBench (74.9--81.4\% vs.\ 81.4\%; \cfgRS{} ties it). External macro accuracy stays within $-1.2$ to $+0.8$ points of the original, while Banking77 gains 4.9--7.7 points and WinoGrande 0.8--4.0.

\paragraph{General ability is preserved.}
No fine-tuned model shows a significant decline in the General average. The 4B model moves from 67.3\% to 67.3--69.0\% and the 0.6B model from 35.6\% to 33.4--36.8\%; the only significant change in an average is a 1.7-point gain for 4B \cfgRL{} (pooled ${p=0.03}$; all others ${p\ge0.08}$). WikiText perplexity remains stable (4B: 11.47 $\rightarrow$ 11.42--11.62; 0.6B: 27.52 $\rightarrow$ 26.31--27.01). The only significant per-task change is a 7.0-point IFEval drop for 0.6B \cfgIL{} (${p=0.04}$; Table~\ref{tab:eval3-full}).

\section{Ablation Details}
\label{app:ablation}

\paragraph{Experimental setup.}
The ablations in \S\ref{sec:recipe} reuse the \cfgLong{} data and schedule (375 steps, 38 warmup steps) and every other setting; only the development health check warns instead of stopping. LoRA adapters wrap every linear layer inside the decoder layers of the language model; embeddings, the LM head, and the vision encoder of Qwen3.5-4B stay frozen. Table~\ref{tab:ablation} gives the results summarized in Figure~\ref{fig:ablation}, Table~\ref{tab:ablation-full} gives per-task results, and Table~\ref{tab:ablation-drift} gives the drift, text-mode, and chat-behavior measurements. Because each configuration is trained with a single random seed, minor fluctuations of one to two percentage points on individual benchmarks (such as Banking77) likely reflect standard run-to-run variance rather than systematic algorithmic divergence.

\paragraph{Aggregate benchmark scores are largely insensitive to anchor strength.}
JevBench accuracy remains largely invariant to regularization strength: relative to ${\lambda=1}$, performance fluctuates by at most 1.7 points at 4B and 3.0 points at 0.6B, with the sole outlier being 4B LoRA under ${\lambda=0.01}$ (79.2\% vs.\ 84.0\%). Without anchors, the listwise loss fits the development data as well as with them (4B tree loss 0.421 vs.\ 0.410) and does not collapse: in every development dataset, legal tokens keep at least 91\% of the probability at the first answer position. General averages do not fall either, and at 4B lighter anchors even raise them (up to $+2.3$ points at ${\lambda=0.01}$, ${p=0.02}$).

\paragraph{Unanchored models exhibit severe generation runaway.}
The listwise loss gives no gradient at the start of the reply or after the answer, so predictions there change only as a side effect of shared weights. Without anchors, the KL divergence from the original model at these two positions grows 44- and 35-fold at 4B, and 51- and 16-fold at 0.6B, relative to ${\lambda=1}$ (Table~\ref{tab:ablation-drift}). In text mode, the 4B model then keeps writing after the identifier on 69 of the 231 JevBench items: 44 answers run to the 32-token limit, mostly with an explanation, and 25 append the option label. When parsing strictly from the leading bracketed identifier, all generated responses agree with the candidate selected during scoring, maintaining an effective text accuracy equal to scoring accuracy (79.7\%). These syntax runaways concentrate predominantly on long-context inputs (35 of the 37 long items). For 0.6B, 176 answers reach the limit, compared with 1 at ${\lambda=1}$. A small weight prevents most of this: at ${\lambda=0.01}$, only one 4B answer continues after the identifier (two with LoRA), while 26 0.6B answers reach the limit.

\paragraph{Unregularized drift compromises open-ended conversational stability.}
The behavior check tests whether this drift extends beyond the decision prompt. At 4B, it does not: under every recipe, including ${\lambda=0}$, about as many chat replies newly fail to end as newly end (19 vs.\ 21 without anchors), and loops stay rare. Trained 4B models only answer short-answer prompts more tersely; at ${\lambda=1}$, for example, the full-parameter model's replies are shorter than the original model's on 103 of the 150 such prompts and longer on 35. At 0.6B, the change grows as the anchor weight falls. With full-parameter training, 7, 13, 12, and 20 replies newly fail to end at ${\lambda=1}$, 0.1, 0.01, and 0, against 1--2 in the other direction; every weight below 1 is significant (${p\le0.013}$). Most come from long-form Dolly prompts and are often endless numbered lists, as in brainstorming answers. The General scores do not reveal this change: the 0.6B model without anchors has a General average 1.0 point higher than the original, because these tasks score likelihoods, extract the final answer, or check only the stated instructions.

\paragraph{Lighter anchors trade behavioral stability for in-domain gains.}
Across both scales, reducing anchor regularization trades conversational stability for marginal in-domain accuracy, confirming ${\lambda=1}$ as the most reliable default (Figure~\ref{fig:ablation}). At 0.6B, no anchor weight produces statistically significant shifts on JevBench relative to ${\lambda=1}$: full-parameter models achieve 57.1--59.3\% and LoRA reaches 58.0--61.0\%, with peak accuracy attained by LoRA at ${\lambda=0.01}$ (61.0\%; 10 wins and 3 losses against LoRA at ${\lambda=1}$, ${p=0.09}$) and optimal calibration delivered by full-parameter tuning at ${\lambda=1}$ (ECE 0.118). While milder regularization lifts Banking77 accuracy under full fine-tuning (${\lambda=0.01}$: 62.4\% vs.\ 60.5\%, ${p<0.001}$; ${\lambda=0.1}$: 62.2\%), all settings below ${\lambda=1}$ induce statistically significant increases in non-terminating chat replies, whereas ${\lambda=1}$ maintains behavioral fidelity closest to the base model (7 newly non-ending vs.\ 1 resolved, ${p=0.07}$). At 4B, the strongest anchor decides best: ${\lambda=1}$ gives the highest JevBench accuracy with both methods and, with full-parameter training, higher Banking77 accuracy than ${\lambda=0}$ (${p=0.002}$), though not significantly higher than ${\lambda=0.1}$ (${p=0.06}$) or ${\lambda=0.01}$. This suggests that the 4B model already decides well and mainly needs the answer format, so departures from the original model add little. Lighter anchors raise full-parameter 4B External accuracy by about 0.7 points (${\lambda=0.1}$: ${p=0.04}$; ${\lambda=0}$: ${p=0.03}$), partly because ${\lambda=1}$ moves ten-option MMLU-Pro answers from option 10 to option 1, whose identifiers share their first token (Appendix~\ref{app:option10}).

\paragraph{Superiority of LoRA on capable backbones.}
LoRA configured with ${\lambda=1}$ achieves 84.0\% on JevBench, marking our strongest 4B model overall (13 wins, 7 losses against the training-free baseline), accompanied by the best calibration of any 4B model without thinking (ECE 0.050 vs.\ 0.057 for the base model; Brier score 0.240 vs.\ 0.247). Three of the four LoRA configurations outperform the training-free baseline (82.3--84.0\%), with only ${\lambda=0.01}$ dropping to 79.2\%. Furthermore, LoRA surpasses full-parameter tuning at ${\lambda=1}$ (+3.9 points; 10 wins vs.\ 1 loss; ${p=0.01}$), maintains advantages at ${\lambda=0.1}$ (+3.0, ${p=0.07}$) and ${\lambda=0}$ (+2.6, ${p=0.15}$), and ties at ${\lambda=0.01}$ (6 wins vs.\ 7 losses); notably, LoRA at ${\lambda=1}$ significantly outperforms every full-parameter ablation run (${p\le0.02}$). At this anchor weight, LoRA matches full fine-tuning on both External (78.9\%) and Banking77 (75.4\% vs.\ 75.2\%). General benchmark averages (66.7\% vs.\ 67.3\%), autoregressive text compliance, and conversational behavior exhibit no significant degradation, altering predictions on only 60 of 1,250 General evaluation items.

\paragraph{Near-parity between LoRA and full fine-tuning at small scale.}
At ${\lambda=1}$, 0.1, and 0, LoRA and full-parameter training are within one JevBench item of each other; at ${\lambda=0.01}$, LoRA reaches 61.0\% against 57.1\% (12 wins, 3 losses; ${p=0.04}$), the best 0.6B result, though not significantly above LoRA at ${\lambda=1}$ (${p=0.09}$). LoRA trails on Banking77 at ${\lambda=1}$ (54.9\% vs.\ 60.5\%) and is 0.3--0.8 points lower on External at every weight, and both methods lose some IFEval accuracy (LoRA 2.0--4.0 points, full-parameter training 3.0--5.0). A higher learning rate ($3\times10^{-4}$) fits the training data slightly better than full-parameter training (development tree loss 0.831 vs.\ 0.838) but does not raise JevBench (58.4\% vs.\ 58.0\% at $10^{-4}$) or External accuracy (Table~\ref{tab:ablation-full}).

\paragraph{Parameter efficiency does not obviate regularization.}
Parameter-efficient training might be expected to protect the original model on its own, but it does not. Without anchors, 9 of the 4B LoRA model's text-mode answers continue after the identifier, mostly with the option label, and 108 of the 0.6B LoRA model's answers reach the 32-token limit. In chat, 0.6B LoRA drifts more than full-parameter training at ${\lambda\le0.01}$: 19 and 38 replies newly fail to end at ${\lambda=0.01}$ and 0, compared with 12 and 20; at ${\lambda=0.1}$ the two are similar (11 vs.\ 13).

\begin{table*}[t]
\centering
\footnotesize
\setlength{\tabcolsep}{4.6pt}
\begin{tabular}{lllrrrrrrr}
\toprule
 & & & \multicolumn{2}{c}{JevBench} & \multicolumn{2}{c}{External} & General & \multicolumn{2}{c}{Behavior} \\
\cmidrule(lr){4-5}\cmidrule(lr){6-7}\cmidrule(lr){8-8}\cmidrule(lr){9-10}
Model & Method & $\lambda$ & Acc. & ECE & Macro & B77 & Avg. & Text & Chat \\
\midrule
\multirow{9}{*}{4B}
 & original & -- & 81.4 & .057 & 78.6 & 69.0 & 67.3 & 0 & -- \\
 & Full & 1 & 80.1 & .070 & 78.9 & 75.2 & 68.8 & 0 & 20/20 \\
 & Full & 0.1 & 79.7 & .086 & 79.6 & 74.4 & 69.0 & 0 & 13/20 \\
 & Full & 0.01 & 79.7 & .091 & 79.0 & 74.9 & \textbf{69.6}$^*$ & 1 & 16/17 \\
 & Full & 0 & 79.7 & .084 & 79.6 & 74.2 & 69.1$^*$ & 69 & 19/21 \\
 & LoRA & 1 & \textbf{84.0} & \textbf{.050} & 78.9 & \textbf{75.4} & 66.7 & 0 & 15/23 \\
 & LoRA & 0.1 & 82.7 & .059 & 79.6 & 74.7 & 67.7 & 0 & 16/22 \\
 & LoRA & 0.01 & 79.2 & .081 & \textbf{79.8} & 74.1 & 68.6 & 2 & 12/22 \\
 & LoRA & 0 & 82.3 & .070 & 79.5 & 73.6 & 68.6 & 9 & 16/23 \\
\midrule
\multirow{9}{*}{0.6B}
 & original & -- & 56.7 & .278 & 57.9 & 22.2 & 35.6 & 0 & -- \\
 & Full & 1 & 58.0 & \textbf{.118} & 59.1 & 60.5 & 35.8 & 1 & 7/1 \\
 & Full & 0.1 & 58.4 & .121 & \textbf{59.2} & 62.2 & 36.1 & 6 & 13/2$^*$ \\
 & Full & 0.01 & 57.1 & .144 & 59.0 & \textbf{62.4} & \textbf{36.6} & 26 & 12/2$^*$ \\
 & Full & 0 & 59.3 & .141 & 58.6 & 61.6 & \textbf{36.6} & 176 & 20/2$^*$ \\
 & LoRA & 1 & 58.0 & .159 & 58.3 & 54.9 & 34.9 & 1 & 7/1 \\
 & LoRA & 0.1 & 58.9 & .150 & 58.9 & 60.2 & 34.6 & 1 & 11/2$^*$ \\
 & LoRA & 0.01 & \textbf{61.0} & .121 & 58.5 & 59.0 & 35.5 & 4 & 19/2$^*$ \\
 & LoRA & 0 & 59.7 & .165 & 57.8 & 59.8 & 35.1 & 108 & 38/2$^*$ \\
\bottomrule
\end{tabular}
\caption{Anchor-weight and LoRA ablations on \cfgLong{} (accuracy in \%). $\lambda$ weights KL anchor penalties (${\lambda=0}$ denotes unconstrained listwise loss). Text reports JevBench text-mode syntax failures (continuations past identifier for 4B; 32-token limits for 0.6B). Chat denotes newly non-terminating vs.\ newly terminating responses across 560 conversational queries. $^*$\,Statistically significant shift from the base model (${p<0.05}$). Table~\ref{tab:ablation-full} gives per-task results.}
\label{tab:ablation}
\end{table*}

\begin{table*}[t]
\centering
\footnotesize
\setlength{\tabcolsep}{3.8pt}
\begin{tabular}{lllrrrrrrrr}
\toprule
Model & Method & $\lambda$ & BoolQ & MMLU & MMLU-Pro & ARC-C & WinoGrande & SciQ & Macro & Banking77 \\
\midrule
\multirow{9}{*}{4B}
 & original & -- & 88.0 & 77.1 & 47.1 & 93.3 & 66.2 & 99.8 & 78.6 & 69.0 \\
 & Full & 1 & 88.0 & 77.2 & 45.9 & 92.0 & 70.2 & 100.0 & 78.9 & 75.2 \\
 & Full & 0.1 & 88.2 & 77.3 & 47.6 & 93.2 & 71.8 & 99.8 & 79.6 & 74.4 \\
 & Full & 0.01 & 87.6 & 76.9 & 46.4 & 93.0 & 70.4 & 99.8 & 79.0 & 74.9 \\
 & Full & 0 & 88.0 & 77.4 & 48.4 & 93.3 & 70.8 & 99.8 & 79.6 & 74.2 \\
 & LoRA & 1 & 88.2 & 77.0 & 46.9 & 92.7 & 68.8 & 100.0 & 78.9 & 75.4 \\
 & LoRA & 0.1 & 88.4 & 77.7 & 48.0 & 93.0 & 70.6 & 100.0 & 79.6 & 74.7 \\
 & LoRA & 0.01 & 88.8 & 78.2 & 47.6 & 92.8 & 71.0 & 100.0 & 79.8 & 74.1 \\
 & LoRA & 0 & 88.0 & 78.2 & 47.1 & 92.3 & 71.2 & 100.0 & 79.5 & 73.6 \\
\midrule
\multirow{10}{*}{0.6B}
 & original & -- & 72.6 & 46.8 & 22.2 & 54.2 & 53.6 & 97.8 & 57.9 & 22.2 \\
 & Full & 1 & 74.4 & 47.8 & 20.5 & 60.3 & 53.6 & 97.8 & 59.1 & 60.5 \\
 & Full & 0.1 & 75.0 & 47.2 & 20.4 & 60.0 & 55.4 & 97.5 & 59.2 & 62.2 \\
 & Full & 0.01 & 74.6 & 47.6 & 20.8 & 59.0 & 55.0 & 97.2 & 59.0 & 62.4 \\
 & Full & 0 & 74.0 & 46.5 & 22.2 & 59.2 & 52.6 & 97.2 & 58.6 & 61.6 \\
 & LoRA & 1 & 72.4 & 46.0 & 20.8 & 57.8 & 55.0 & 97.8 & 58.3 & 54.9 \\
 & LoRA & 0.1 & 71.6 & 47.5 & 22.0 & 58.7 & 56.0 & 97.8 & 58.9 & 60.2 \\
 & LoRA & 0.01 & 72.8 & 47.0 & 21.5 & 57.2 & 55.2 & 97.5 & 58.5 & 59.0 \\
 & LoRA & 0 & 71.4 & 46.0 & 20.6 & 56.5 & 54.6 & 97.8 & 57.8 & 59.8 \\
 & LoRA, LR $3\times10^{-4}$ & 1 & 69.6 & 46.4 & 21.0 & 57.7 & 54.8 & 98.0 & 57.9 & 57.9 \\
\bottomrule
\end{tabular}

\vspace{8pt}
\begin{tabular}{lllrrrrrr}
\toprule
Model & Method & $\lambda$ & GSM8K & IFEval & TriviaQA & LAMBADA & Avg. & WikiText ppl \\
\midrule
\multirow{9}{*}{4B}
 & original & -- & 84.0 & 80.5 & 41.3 & 63.2 & 67.3 & 11.47 \\
 & Full & 1 & 87.2 & 82.0 & 42.3 & 63.6 & 68.8 & 11.44 \\
 & Full & 0.1 & 85.6 & 85.0 & 42.3 & 63.2 & 69.0 & 11.43 \\
 & Full & 0.01 & 88.0 & 84.0 & 43.0 & 63.4 & 69.6$^*$ & 11.41 \\
 & Full & 0 & 87.2 & 82.0 & 43.0 & 64.2 & 69.1$^*$ & 11.41 \\
 & LoRA & 1 & 83.2 & 79.5 & 41.3 & 62.8 & 66.7 & 11.41 \\
 & LoRA & 0.1 & 86.8 & 78.5 & 42.0 & 63.6 & 67.7 & 11.42 \\
 & LoRA & 0.01 & 86.8 & 81.0 & 43.3 & 63.2 & 68.6 & 11.43 \\
 & LoRA & 0 & 87.2 & 80.5 & 42.0 & 64.8 & 68.6 & 11.40 \\
\midrule
\multirow{10}{*}{0.6B}
 & original & -- & 44.4 & 55.0 & 4.0 & 39.0 & 35.6 & 27.52 \\
 & Full & 1 & 48.4 & 51.0 & 4.0 & 39.8 & 35.8 & 26.64 \\
 & Full & 0.1 & 48.8 & 52.0 & 2.3 & 41.2 & 36.1 & 26.50 \\
 & Full & 0.01 & 50.8$^*$ & 51.5 & 3.0 & 41.2 & 36.6 & 26.32 \\
 & Full & 0 & 54.0$^*$ & 50.0 & 2.3 & 40.2 & 36.6 & 26.30 \\
 & LoRA & 1 & 44.0 & 53.0 & 3.7 & 38.8 & 34.9 & 26.19 \\
 & LoRA & 0.1 & 44.0 & 51.5 & 2.3 & 40.6 & 34.6 & 26.37 \\
 & LoRA & 0.01 & 47.2 & 52.0 & 2.3 & 40.6 & 35.5 & 26.22 \\
 & LoRA & 0 & 48.4 & 51.0 & 1.7$^*$ & 39.4 & 35.1 & 26.28 \\
 & LoRA, LR $3\times10^{-4}$ & 1 & 45.2 & 53.5 & 3.0 & 38.6 & 35.1 & 26.75 \\
\bottomrule
\end{tabular}
\caption{Per-task breakdown of ablation configurations: External (top) and General (bottom) benchmarks (accuracy in \%; lower perplexity is better). $^*$\,Statistically significant shift from the base model (${p<0.05}$).}
\label{tab:ablation-full}
\end{table*}

\begin{table*}[t]
\centering
\footnotesize
\setlength{\tabcolsep}{5pt}
\begin{tabular}{lllrrrrr}
\toprule
 & & & \multicolumn{3}{c}{Drift (development sets)} & \multicolumn{2}{c}{JevBench text mode} \\
\cmidrule(lr){4-6}\cmidrule(lr){7-8}
Model & Method & $\lambda$ & KL start & KL answer & Tree loss & After \texttt{[$k$]} & At limit \\
\midrule
\multirow{9}{*}{4B}
 & original & -- & 0 & 0 & 0.763 & 0 & 0 \\
 & Full & 1 & .0011 & .0088 & 0.410 & 0 & 0 \\
 & Full & 0.1 & .0022 & .017 & 0.417 & 0 & 0 \\
 & Full & 0.01 & .0033 & .029 & 0.416 & 1 & 0 \\
 & Full & 0 & .048 & .305 & 0.421 & 69 & 44 \\
 & LoRA & 1 & .0007 & .0045 & 0.433 & 0 & 0 \\
 & LoRA & 0.1 & .0020 & .014 & 0.424 & 0 & 0 \\
 & LoRA & 0.01 & .0057 & .029 & 0.424 & 2 & 0 \\
 & LoRA & 0 & .222 & .114 & 0.433 & 9 & 1 \\
\midrule
\multirow{10}{*}{0.6B}
 & original & -- & 0 & 0 & 3.158 & 219 & 0 \\
 & Full & 1 & .062 & .076 & 0.838 & 230 & 1 \\
 & Full & 0.1 & .146 & .109 & 0.833 & 221 & 6 \\
 & Full & 0.01 & .294 & .179 & 0.832 & 216 & 26 \\
 & Full & 0 & 3.19 & 1.18 & 0.822 & 231 & 176 \\
 & LoRA & 1 & .050 & .052 & 0.921 & 223 & 1 \\
 & LoRA & 0.1 & .138 & .127 & 0.847 & 216 & 1 \\
 & LoRA & 0.01 & .261 & .185 & 0.867 & 213 & 4 \\
 & LoRA & 0 & 2.17 & .753 & 0.877 & 231 & 108 \\
 & LoRA, LR $3\times10^{-4}$ & 1 & .065 & .067 & 0.831 & 227 & 1 \\
\bottomrule
\end{tabular}

\vspace{8pt}
\begin{tabular}{lllrrrrr}
\toprule
 & & & \multicolumn{5}{c}{Chat (560 replies)} \\
\cmidrule(lr){4-8}
Model & Method & $\lambda$ & No end & New/res. & Loops & New/res. & Shorter/longer \\
\midrule
\multirow{9}{*}{4B}
 & original & -- & 69 & -- & 1 & -- & -- \\
 & Full & 1 & 69 & 20/20 & 3 & 3/1 & 103/35$^*$ \\
 & Full & 0.1 & 62 & 13/20 & 1 & 1/1 & 105/34$^*$ \\
 & Full & 0.01 & 68 & 16/17 & 2 & 2/1 & 112/30$^*$ \\
 & Full & 0 & 67 & 19/21 & 2 & 2/1 & 105/36$^*$ \\
 & LoRA & 1 & 61 & 15/23 & 1 & 1/1 & 90/43$^*$ \\
 & LoRA & 0.1 & 63 & 16/22 & 2 & 2/1 & 90/45$^*$ \\
 & LoRA & 0.01 & 59 & 12/22 & 2 & 2/1 & 115/24$^*$ \\
 & LoRA & 0 & 62 & 16/23 & 1 & 1/1 & 100/41$^*$ \\
\midrule
\multirow{10}{*}{0.6B}
 & original & -- & 3 & -- & 6 & -- & -- \\
 & Full & 1 & 9 & 7/1 & 12 & 11/5 & 85/40$^*$ \\
 & Full & 0.1 & 14 & 13/2$^*$ & 13 & 12/5 & 67/61 \\
 & Full & 0.01 & 13 & 12/2$^*$ & 13 & 13/6 & 75/61 \\
 & Full & 0 & 21 & 20/2$^*$ & 16 & 14/4$^*$ & 67/71 \\
 & LoRA & 1 & 9 & 7/1 & 13 & 11/4 & 60/58 \\
 & LoRA & 0.1 & 12 & 11/2$^*$ & 11 & 9/4 & 61/66 \\
 & LoRA & 0.01 & 20 & 19/2$^*$ & 20 & 19/5$^*$ & 84/47$^*$ \\
 & LoRA & 0 & 39 & 38/2$^*$ & 33 & 32/5$^*$ & 59/78 \\
 & LoRA, LR $3\times10^{-4}$ & 1 & 11 & 10/2$^*$ & 12 & 10/4 & 83/47$^*$ \\
\bottomrule
\end{tabular}
\caption{Auxiliary distribution drift and conversational behavior across ablations. KL start and KL answer denote mean KL divergence against base models at the reply prefix and post-bracket positions across 12 validation sets. Text mode counts syntax runaways out of 231 JevBench queries. Chat reports non-terminating (No end) and looping responses on MT-Bench and Dolly-15k. $^*$\,${p<0.05}$.}
\label{tab:ablation-drift}
\end{table*}

\paragraph{Option 1 versus option 10.}
\label{app:option10}
MMLU-Pro is the only ranked External benchmark with two-digit identifiers: 658 of its 800 questions have ten options, and the suffixes of options 1 and 10 share their first token. Training shifts answers away from option 10, more so at larger anchor weights. With full-parameter training, the 4B model picks option 10 on 45 of these questions before training and on 31, 32, 25, and 24 at ${\lambda=0}$, 0.01, 0.1, and 1; the 0.6B model does so on 18 before training and on 16, 9, 7, and 2. At 4B with ${\lambda=1}$, the displaced answers go to option 1 (147 picks vs.\ 123), and the model loses 15 of these questions net, more than its total net MMLU-Pro loss of 10. Investigating the precise representation dynamics behind this prefix-collision effect remains an interesting direction for future work.

\section{Anchor Sites and Sizes}
\label{app:anchor-sites}

\paragraph{Setup.}
To test whether each anchor site is needed, we retrain \cfgLong{} with ${\lambda=1}$ and all parameters at both scales, removing one part of the anchors at a time (Table~\ref{tab:anchor-sites}): the two context positions, the off-path nodes (so that only the gold path is anchored), or the two distribution anchors $\mathcal{L}_{\text{out}}$ and $\mathcal{L}_{\text{pos}}$, which leaves the legal mass alone and is equivalent to keeping ${k=0}$ tokens. To size the truncation and the number of off-path nodes without training more models, we compare the original model with the trained models on the first 64 development examples of each of the 12 training datasets (the \emph{analysis examples}), at the sites that training anchors; there we sample eight off-path nodes per example instead of three.

\begin{table*}[t]
\centering
\footnotesize
\setlength{\tabcolsep}{4.6pt}
\begin{tabular}{llrrrrrrrrr}
\toprule
 & & \multicolumn{2}{c}{JevBench} & \multicolumn{2}{c}{External} & General & \multicolumn{2}{c}{Behavior} & \multicolumn{2}{c}{Drift} \\
\cmidrule(lr){3-4}\cmidrule(lr){5-6}\cmidrule(lr){7-7}\cmidrule(lr){8-9}\cmidrule(lr){10-11}
Model & Anchors & Acc. & ECE & Macro & B77 & Avg. & Text & Chat & KL start & KL answer \\
\midrule
\multirow{5}{*}{4B}
 & all & 80.1 & .070 & 78.9 & 75.2 & 68.8 & 0 & 20/20 & .0011 & .0088 \\
 & no context positions & 80.1 & .089 & 79.3 & 75.8$^\dagger$ & 68.0 & 3 & 15/18 & .010 & .021 \\
 & no off-path nodes & 78.8 & .071 & 78.6 & 73.4$^\dagger$ & 68.4 & 0 & 14/17 & .0012 & .010 \\
 & legal mass only (${k=0}$) & 80.5 & .070 & 79.3 & 74.8 & 69.0 & 16 & 26/18 & .104 & .136 \\
 & none (${\lambda=0}$) & 79.7 & .084 & 79.6$^\dagger$ & 74.2$^\dagger$ & 69.1 & 69 & 19/21 & .048 & .305 \\
\midrule
\multirow{5}{*}{0.6B}
 & all & 58.0 & .118 & 59.1 & 60.5 & 35.8 & 1 & 7/1 & .062 & .076 \\
 & no context positions & 57.6 & .143 & 58.7 & 60.8 & 35.6 & 3 & 9/2 & .786 & .199 \\
 & no off-path nodes & 57.1 & .165 & 58.8 & 61.0 & 35.2 & 1 & 12/1$^*$ & .066 & .077 \\
 & legal mass only (${k=0}$) & 57.6 & .132 & 59.0 & 62.0$^\dagger$ & 37.2 & 39 & 14/2$^*$ & 2.52 & .860 \\
 & none (${\lambda=0}$) & 59.3 & .141 & 58.6 & 61.6$^\dagger$ & 36.6 & 176 & 20/2$^*$ & 3.19 & 1.18 \\
\bottomrule
\end{tabular}
\caption{Anchor-site ablations on \cfgLong{} under full-parameter training (accuracy in \%; ${\lambda=1}$). Ablations omit context positions ($\mathcal{L}_{\text{pos}}$), off-path nodes, or both distribution anchors ($\mathcal{L}_{\text{out}}$, $\mathcal{L}_{\text{pos}}$; equivalent to ${k=0}$). $^\dagger$\,Significant shift from all anchors (${p<0.05}$). $^*$\,Significant conversational shift from base model.}
\label{tab:anchor-sites}
\end{table*}

\paragraph{Every anchor site serves a distinct regularizing role.}
While no single site ablation significantly degrades JevBench accuracy (${p\ge0.38}$), each omission directly destabilizes the specific distribution it targets. Omitting context positions inflates KL divergence at the reply start by 9.4$\times$ at 4B and 12.6$\times$ at 0.6B (and 2.4$\times$ and 2.6$\times$ after the bracket); consequently, three text-mode responses emit runaway text at 4B and three reach the length limit at 0.6B, while JevBench ECE rises from 0.070 to 0.089 at 4B and from 0.118 to 0.143 at 0.6B. Removing off-path node anchors increases KL divergence over illegal tokens at unvisited branches by 2.9$\times$ at 4B (0.084 to 0.241) and 1.8$\times$ at 0.6B (0.176 to 0.319), elevating 0.6B ECE to 0.165. Constraining legal mass alone fails to prevent generative runaway at both scales (16 4B completions append trailing text and 39 0.6B completions hit the length cap), causing reply-start KL to surge 95-fold at 4B and 40-fold at 0.6B---comparable to unconstrained training without any anchors (44- and 51-fold). In conversational chat, omitting off-path nodes or isolating legal mass induces significantly more non-terminating replies than the base model at 0.6B. Variations on Banking77 remain bounded within $\pm1.8$ points; given single-seed runs, we do not attribute these modest shifts to individual anchor components.

\paragraph{Top-64 truncation captures the distribution tail.}
At candidate trie nodes, the base model's 64 most probable illegal tokens account on average for 94.5--98.9\% of total illegal probability mass (rising to 97.1--99.6\% with 1,024 tokens), while at context positions the top-64 tokens cover over 99.9\% of next-token probability. By the chain rule of KL divergence, the true divergence exceeds our truncated, bucketed objective only by the product of the base model's tail mass and the residual divergence within the tail; the bound is therefore exceptionally tight when tail mass is light. Across anchored models, this truncation gap averages at most $7.3\times10^{-4}$ nats per anchored site ($1.9\times10^{-3}$ without anchors). Moreover, caching truncated distributions consumes only $\approx$0.5\,KB per distribution, compared to 0.6\,MB (Qwen3-0.6B) or 1.0\,MB (Qwen3.5-4B) for the full vocabulary in FP32. Truncating aggressively to ${k=0}$ completely removes distributional constraints, reproducing the degenerate behavior observed in legal-mass-only runs.

\paragraph{Sampling three off-path nodes is computationally optimal.}
Because a four-option query contains exactly three off-path candidate nodes, setting the sample size to three provides exhaustive off-path coverage for 9 of the 12 training datasets in \cfgLong{}; only the three intent datasets (spanning 27--151 options) require stochastic subsampling across iterations. Regularizing three randomly sampled nodes per training instance maintains all off-path branches close to the base model: on held-out analysis instances where eight off-path nodes are evaluated per intent query, off-path KL divergence (0.084 at 4B, 0.176 at 0.6B) remains virtually identical to gold-path divergence (0.089 and 0.192), where every node is anchored deterministically. Increasing the number of sampled nodes yields diminishing returns while imposing substantial computational overhead, as each anchored node requires an additional forward evaluation over the prompt prefix. Sampling three nodes increases evaluated forward tokens 3.4-fold per instance, resulting in a $2.25\times$ slowdown at 4B and $2.23\times$ at 0.6B relative to unanchored training.

\section{Community Jev-Style Models}
\label{app:community}

Table~\ref{tab:community} summarizes the community Jev-style models most closely related to our work on JevBench v1.4.2.2. Information comes from each system's README or model card and the evaluator's notes, accessed on 2026-09-28 (LLM2Jev: 2026-10-04). ``Public'' reports official accuracy on the 231 public items. ``Sealed'' reports accuracy on 308 unreleased items, where random guessing achieves about 29.3\%.

\begin{table*}[t]
\centering
\footnotesize
\setlength{\tabcolsep}{3pt}
\renewcommand{\arraystretch}{1.12}
\begin{tabular}{>{\raggedright\arraybackslash}p{2.9cm}>{\raggedright\arraybackslash}p{2.4cm}>{\raggedright\arraybackslash}p{5.5cm}>{\raggedright\arraybackslash}p{2.3cm}rr}
\toprule
System & Base & Training & Readout & Public & Sealed \\
\midrule
\mbox{Imajev-4B} \citep{imajev} & Qwen3.5-4B & LoRA in four stages (human data, pseudo-labels, teacher questions, soft labels), weight-averaged & new head + temperature & 86.1 & 37.0 \\
\mbox{Plumb-4B} \citep{plumb} & JevK5 v0.2 & LoRA over five rounds: teacher questions, mined errors, long documents, replayed public data & letter logits + temperature & 89.6 & 38.0 \\
\mbox{decider-4b v2} \citep{decider} & Qwen3.5-4B-Base & full SFT, then LoRA; about 95 public sets, programmatic and teacher data & letter logits + temperature & 83.5 & 34.7 \\
\mbox{Jev 1.13.0} \citep{typesafe2026jev} & undisclosed & RLCD (details undisclosed) & -- & 86.6 & 36.7 \\
\mbox{JevK5 v0.2} \citep{jevk5} & Qwen3.5-4B & distilled LoRA: 3,272 teacher questions plus equal public replay & letter logits + temperature & 85.3 & 33.1 \\
Cygnet \citep{cygnet} & Gemma-4-12B-it & none & letter logits + temperature & 87.9 & 33.8 \\
Hopper \citep{hopper} & Qwen3.5-4B & LoRA; synthetic families + public data & letter logits + per-type temperature & 82.3 & 34.1 \\
\mbox{Winnow-12B} \citep{winnow} & Gemma-4-12B-it & LoRA; private synthetic and teacher data & answer-token logits & 85.7 & 33.1 \\
\mbox{reflex 4B} \citep{reflex} & Qwen3.5-4B & LoRA (the authors now ship the frozen model) & label logits + temperature & 79.2 & 28.2 \\
SemIf \citep{lee2026semif} & Qwen3.5-4B & none & letter logits & 81.0 & 26.3 \\
\mbox{open-alternative-jev} \citep{openalternativejev} & Qwen3.5-4B & none & letter logits + temperature & 74.0 & 24.4 \\
LLM2Jev \citep{yinsongxu2026llm2jev} & Qwen3.5-4B & none & yes/no logits, one prompt per option & 76.2$^\S$ & -- \\
jqv \citep{jqv} & Qwen3-32B & none & letter logits + temperature & 80.1 & 28.2 \\
\mbox{Kev-0.6B} \citep{kev} & Qwen3-0.6B-Base & LoRA + pointer head; public and programmatic data & new head & 66.7 & 24.0 \\
\mbox{Decision Fast} \citep{decisionfast} & Qwen3-0.6B-Base & LoRA + pointer head & new head + temperature & 63.2 & 25.6 \\
\midrule
\sys{} 4B, training-free (ours) & Qwen3.5-4B & none & multi-token suffix & 81.4$^\ddagger$ & -- \\
\sys{} 4B, LoRA (ours) & Qwen3.5-4B & LoRA + KL anchors; public data & multi-token suffix & 84.0$^\ddagger$ & -- \\
\sys{} 0.6B, \cfgLong{} (ours) & Qwen3-0.6B & full parameters + KL anchors; public data & multi-token suffix & 58.0$^\ddagger$ & -- \\
\bottomrule
\end{tabular}
\caption{Community Jev-style models on JevBench v1.4.2.2. $^\ddagger$\,Public-item accuracy from our evaluation; local reruns reproduce official standings within four items. $^\S$\,Reported by the project from its own run on the public items.}
\label{tab:community}
\end{table*}

\section{Budgeted Thinking Details}
\label{app:thinking}

\paragraph{Prompts.}
Each prompt below adds a sentence on a new line after the user message of Appendix~\ref{app:impl}, with $N$ replaced by the budget:
{\raggedright
\begin{itemize}
\item \emph{Limit}: \texttt{Think briefly before you answer: your thinking is limited to $N$ tokens.} The answer prefix stays \texttt{Best answer: [}.
\item \emph{Limit + answer}: the same sentence followed by \texttt{At the end of your thinking, you must provide the best answer.} The answer prefix is \texttt{After thinking, the best answer is [}.
\item \emph{Limit + answer + stop}: as \emph{Limit + answer}, and a thought cut off at the budget gets the stop-thinking sentence of Qwen's thinking budget~\citep{yang2025qwen3}, \texttt{Considering the limited time by the user, I have to give the solution based on the thinking directly now.}, as its last paragraph.
\end{itemize}
\par}

\paragraph{Thinking and scoring.}
We apply the chat template with \texttt{enable\_thinking=True}, which ends the prompt with \texttt{<think>} and a line break. Thinking uses greedy decoding and stops at the model's own \texttt{</think>}, which counts toward the budget, at its end-of-turn token, which never occurred, or after $N$ tokens. We strip whitespace around the thought and close it with a line break, \texttt{</think>}, and two line breaks, the string the chat template places after a finished thought; for an empty thought, this reproduces the template's prompt with thinking disabled. The answer prefix follows, and the candidate suffixes are scored as in Appendix~\ref{app:impl}. Before each of the 19 runs on the RTX A5500, we score three items both with cached candidate branches and by full recomputation after thinking; the two agree to within $4\times10^{-5}$ in probability in every run. In Transformers, Qwen3.5's generation leaves rotary-position offsets on the model, so we restore them after thinking; cached scoring fails otherwise.

\paragraph{Runs.}
Each configuration runs once, in FP32 on one RTX A5500, like the other JevBench results. The supplementary 4,096-token run used an RTX 6000 Ada GPU; on it, the model without thinking chooses the same option as on the RTX A5500 for all 231 items, with probabilities within $6\times10^{-4}$, but its latency is not comparable. Table~\ref{tab:thinking-full} reports every full run we made; no prompt or budget was selected on JevBench.

\begin{table*}[t]
\centering
\footnotesize
\setlength{\tabcolsep}{4.2pt}
\begin{tabular}{lrrrrrrrrrrr}
\toprule
Prompt & Budget & Acc. & Won/lost ($p$) & Easy & Std. & Hard & ECE & Brier & Tokens & Finished & p50 (s) \\
\midrule
No thinking & -- & 81.4 & -- & 100.0 & 94.4 & 64.9 & .057 & .247 & -- & -- & 0.2 \\
\midrule
\multirow{8}{*}{Limit}
 & 32 & 81.0 & 6/7 (1.0) & 100.0 & 95.8 & 63.1 & .052 & .252 & 32 & 0 & 1.3 \\
 & 64 & 81.4 & 9/9 (1.0) & 100.0 & 94.4 & 64.9 & .042 & .247 & 64 & 0 & 2.2 \\
 & 128 & 82.7 & 9/6 (.61) & 100.0 & 97.2 & 65.8 & .059 & .244 & 128 & 0 & 4.2 \\
 & 256 & 80.5 & 9/11 (.82) & 100.0 & 98.6 & 60.4 & .075 & .247 & 255 & 9 & 8.0 \\
 & 512 & 84.8 & 17/9 (.17) & 100.0 & 100.0 & 68.5 & .072 & .198 & 462 & 104 & 15.6 \\
 & 1,024 & 89.2 & 24/6 (.001) & 100.0 & 98.6 & 78.4 & .042 & .158 & 711 & 133 & 19.6 \\
 & 2,048 & 91.8 & 28/4 ($<$.001) & 100.0 & 98.6 & 83.8 & .066 & .136 & 1,115 & 162 & 22.4 \\
 & 4,096$^\dagger$ & 94.8 & 35/4 ($<$.001) & 100.0 & 98.6 & 90.1 & .042 & .078 & 1,822 & 172 & 27.5$^\dagger$ \\
\midrule
\multirow{6}{*}{Limit + answer}
 & 32 & 79.7 & 4/8 (.39) & 100.0 & 93.1 & 62.2 & .057 & .268 & 32 & 0 & 1.3 \\
 & 64 & 81.0 & 8/9 (1.0) & 100.0 & 94.4 & 64.0 & .044 & .261 & 64 & 0 & 2.2 \\
 & 128 & 80.1 & 7/10 (.63) & 100.0 & 98.6 & 59.5 & .076 & .249 & 128 & 0 & 4.1 \\
 & 256 & 81.0 & 10/11 (1.0) & 100.0 & 98.6 & 61.3 & .060 & .249 & 256 & 6 & 8.0 \\
 & 512 & 85.3 & 19/10 (.14) & 100.0 & 100.0 & 69.4 & .051 & .197 & 481 & 76 & 15.7 \\
 & 1,024 & 89.2 & 25/7 (.002) & 100.0 & 98.6 & 78.4 & .040 & .144 & 761 & 115 & 31.0 \\
\midrule
\multirow{6}{*}{Limit + answer + stop}
 & 32 & 78.4 & 5/12 (.14) & 100.0 & 93.1 & 59.5 & .061 & .272 & 32 & 0 & 1.3 \\
 & 64 & 81.4 & 8/8 (1.0) & 100.0 & 95.8 & 64.0 & .045 & .261 & 64 & 0 & 2.3 \\
 & 128 & 79.7 & 7/11 (.48) & 100.0 & 95.8 & 60.4 & .079 & .249 & 128 & 0 & 4.2 \\
 & 256 & 80.1 & 9/12 (.66) & 100.0 & 97.2 & 60.4 & .055 & .256 & 256 & 6 & 8.0 \\
 & 512 & 84.8 & 18/10 (.18) & 100.0 & 98.6 & 69.4 & .066 & .201 & 481 & 76 & 15.7 \\
 & 1,024 & 89.6 & 23/4 ($<$.001) & 100.0 & 98.6 & 79.3 & .040 & .138 & 761 & 115 & 31.2 \\
\bottomrule
\end{tabular}
\caption{Budgeted thinking with the unmodified Qwen3.5-4B on the 231 public JevBench items (accuracy in \%). Won/lost pairs and two-sided sign-test $p$-values compare each run with the same model without thinking. Tokens: mean thinking tokens per decision, counting a \texttt{</think>} written by the model; Finished: thoughts the model ended within the budget; p50: median latency per decision in FP32 on one RTX A5500. $^\dagger$\,Supplementary run on an RTX 6000 Ada GPU, whose latency is not comparable.}
\label{tab:thinking-full}
\end{table*}

\paragraph{What the thoughts contain.}
The preamble described in \S\ref{sec:thinking} opens 227 of 231 thoughts at 32 tokens, 224 at 64, and 206 at 128. With \emph{Limit}, finished thoughts take a median of about 390 tokens on easy items and 470--510 on standard items at budgets from 1,024 to 4,096 tokens, and 1,238 and 1,744 tokens on hard items at budgets of 2,048 and 4,096. Temporal-numeric questions remain the hardest: the model answers 5 of 15 correctly without thinking, 7 at 2,048 tokens, and 9 at 4,096. Of the 12 errors at 4,096 tokens, 11 come from thoughts cut off at the budget, and 6 of these are temporal-numeric items. At the 4,096-token budget, three thoughts (two hard, one easy) exhibit repetitive verification loops (e.g., repeating \texttt{Wait, I need to check...}) until reaching the budget cap; all three are nonetheless correctly answered by the subsequent readout.

\section{Inference Cost and Latency Details}
\label{app:cost}

\begin{table*}[t]
\centering
\footnotesize
\setlength{\tabcolsep}{3.6pt}
\begin{tabular}{lrrrrrrrrr}
\toprule
 & \multicolumn{2}{c}{JevBench acc.} & \multicolumn{2}{c}{\$ / MTok} & \multicolumn{5}{c}{Per decision} \\
\cmidrule(lr){2-3}\cmidrule(lr){4-5}\cmidrule(lr){6-10}
System & All & Hard & Input & Output & In tokens & Out tokens & TFLOPs & Passes & \$ / 1M dec. \\
\midrule
GPT-5.6 Sol (generation) & 94.4 & 88.3 & 4.00 & 20.00 & 584 & 7 & -- & 7 & 2,477 \\
Existing LLM APIs$^\dagger$ & -- & -- & 0.20--10 & $\sim$5$\times$ & 584 & 7 & -- & -- & 124--6,191 \\
Jev$^\dagger$ & 86.6$^\ast$ & -- & 0.042 & free & 653 & -- & -- & -- & 27 \\
\midrule
\multicolumn{10}{l}{\emph{\sys{} (ours)}} \\
4B readout & 81.4 & 64.9 & 0.046 & free & 653 & 0 & 4.8 & 2 & 30 \\
\quad + thinking, 512 tokens & 84.8 & 68.5 & 0.046 & 0.23 & 672 & 462 & 8.9 & 464 & 137 \\
\quad + thinking, 1,024 tokens & 89.2 & 78.4 & 0.046 & 0.23 & 673 & 711 & 11.1 & 713 & 194 \\
\quad + thinking, 2,048 tokens & 91.8 & 83.8 & 0.046 & 0.23 & 673 & 1,115 & 14.7 & 1,117 & 287 \\
\quad + thinking, 4,096 tokens$^\ddagger$ & 94.8 & 90.1 & 0.046 & 0.23 & 673 & 1,822 & 21.1 & 1,824 & 450 \\
4B LoRA & 84.0 & 68.5 & 0.046 & free & 653 & 0 & 4.8 & 2 & 30 \\
0.6B readout & 56.7 & 33.3 & 0.007 & free & 640 & 0 & 0.7 & 2 & 4 \\
\bottomrule
\end{tabular}
\caption{Accuracy on public JevBench (\%) and estimated inference cost per decision. TFLOPs and sequential forward passes (Passes) do not depend on the serving stack. Prices for our models scale the \$0.20/MTok of a 16B dense model by FLOPs per token, with output at five times input; GPT-5.6 Sol is at its promotional API price and generates its answer. Tokens are means over the 231 items, counted with Qwen3.5's tokenizer for Jev and our models and taken from the API's usage reports for GPT-5.6 Sol and the API range; dollars per million decisions multiply them by the prices. $^\dagger$\,Prices claimed by \citet{typesafe2026jev}, with output about five times input for existing LLMs. $^\ast$\,Official leaderboard score (Table~\ref{tab:community}). $^\ddagger$\,Supplementary run (Appendix~\ref{app:thinking}).}
\label{tab:cost}
\end{table*}

Measured latency and price depend on the serving stack: hardware, precision, batching, and caching. We therefore count quantities that do not, the floating-point operations (FLOPs) and the sequential forward passes per decision, and we convert FLOPs into prices with models whose price and size are both public (Table~\ref{tab:cost}).

\paragraph{FLOPs per token.}
A forward pass costs about two FLOPs per parameter and token~\citep{kaplan2020scaling}. For each text token, Qwen3.5-4B runs its 3.57B decoder parameters; its tied embedding matrix acts as the output layer only where logits are needed, and its vision encoder and multi-token-prediction head stay idle. At JevBench's prompt lengths, attention in its 8 full-attention layers and the state updates of its 24 linear-attention layers add 3\%, for 7.36 GFLOPs per input token. Qwen3-0.6B spends 0.88 GFLOPs in its linear layers and 0.23 GFLOPs in attention at JevBench's lengths, plus 0.31 GFLOPs in its output layer at each position with logits.

\paragraph{From FLOPs to price.}
Jev's announcement prices the input of existing LLMs at \$0.20 to \$10 per million tokens (MTok), with output about five times more expensive, and Jev's input at \$0.042/MTok with free output~\citep{typesafe2026jev}. Open-weight models are sold today at the low end of this range. Fireworks charges \$0.20/MTok for any dense model with 4B--16B parameters that it does not list individually, the largest of which need 32 GFLOPs per token~\citep{fireworks2026pricing}, and Together AI charges \$0.20/MTok of input for Qwen3-235B-A22B, which activates 22B parameters, or 44 GFLOPs per token~\citep{together2026pricing}. At the same price per FLOP, Qwen3.5-4B's input would cost \$0.046/MTok against the dense reference and \$0.034/MTok against the sparse one, which brackets Jev's \$0.042. In fact, Together AI already sells a decision model fine-tuned from Qwen3.5-4B, Tev1 4B Experimental, at \$0.04/MTok of input with free output~\citep{together2026tev1}. We use the more conservative \$0.046/MTok below.

\paragraph{The readout's output is nearly free.}
Like Jev, the readout generates no text. After the prompt's prefill, one batched step scores the closing bracket of every candidate suffix (JevBench items have at most six options), which adds 2.3\% to the prompt's FLOPs on average. A decision thus costs about as much as its input: \$30 per million JevBench decisions, against \$27 at Jev's price and \$2,477 for GPT-5.6 Sol at its API price on 2026-10-12, a promotional price offered at least through 2026-11-21~\citep{openai2026gpt56sol}. Fine-tuning does not change these costs, since the fine-tuned models keep the architecture and LoRA weights are merged.

\paragraph{Thinking is priced as output.}
Thinking tokens are decoded one at a time, a regime limited by memory bandwidth rather than compute~\citep{pope2023efficiently}, which is why APIs charge more for output; we price them at five times the input price, the ratio Jev's announcement gives for existing LLMs. At budgets of 1,024 to 4,096 tokens, thinking multiplies the FLOPs of a decision by 2.3 to 4.4 but its price by 6.5 to 15, since its 711 to 1,822 thinking tokens on average are billed as output. It also replaces the readout's two sequential forward passes with 713 to 1,824, so its latency grows with the thought on any hardware.

\paragraph{When to think.}
With a 4,096-token budget, the 4B model reaches GPT-5.6 Sol's accuracy (94.8\% vs.\ 94.4\%) at an estimated \$450 per million decisions, 5.5 times less than GPT-5.6 Sol, which we query without reasoning. Thinking Jev therefore empowers decisions where accuracy is paramount (deliberative System~2), while the default readout preserves a Jev-like price (\$30 per million decisions) and two sequential forward passes for high-volume, real-time System~1 pipelines.

\paragraph{Sensitivity.}
The price per FLOP varies across providers and models. Against the sparse reference, all of our prices fall by 27\%, to \$22 per million decisions for the readout and \$327 at a 4,096-token budget. Pricing output at four times input, as both providers do for gpt-oss-120B, lowers the cost of thinking to \$162 at 1,024 tokens and \$366 at 4,096. Size-tiered list prices charge small models more per FLOP: at Fireworks' \$0.10 or \$0.20/MTok for models below or above 4B parameters, the readout would cost \$65 or \$131 per million decisions. The latencies we measured (Appendix~\ref{app:impl}) come from an unoptimized FP32 implementation and do not enter these estimates.

\end{document}